\documentclass{article}
\usepackage{arxiv}
\usepackage{amssymb}
\usepackage{amsmath}
\usepackage{booktabs}
\usepackage{algorithm}
\usepackage[noend]{algpseudocode}
\usepackage{multirow}
\usepackage{subcaption}
\usepackage{siunitx}
\usepackage{hyperref}
\usepackage{xcolor} 
\usepackage{graphicx} 
\usepackage{orcidlink}

\usepackage[authoryear]{natbib}
\title{RAPID: Risk of Attribute Prediction-Induced Disclosure in Synthetic Microdata}

\author{
Matthias Templ\orcidlink{0000-0002-8638-5276}\thanks{Corresponding author: matthias.templ@fhnw.ch}%
\thanks{School of Business, University of Applied Sciences and Arts Northwestern Switzerland (FHNW), Olten, Switzerland}
\and
Oscar Thees\orcidlink{0009-0001-9378-4988}\footnotemark[2]
\and
Roman M\"uller\orcidlink{0009-0007-2142-3896}\footnotemark[2]
}

\date{}

\begin{document}

\maketitle


\begin{abstract}
Statistical data anonymization increasingly relies on fully synthetic microdata, for which classical identity disclosure measures are less informative than an adversary's ability to infer sensitive attributes from released data. We introduce RAPID (Risk of Attribute Prediction--Induced Disclosure), a disclosure risk measure that directly quantifies inferential vulnerability under a realistic attack model. An adversary trains a predictive model solely on the released synthetic data and applies it to real individuals' quasi-identifiers. For continuous sensitive attributes, RAPID reports the proportion of records whose predicted values fall within a specified relative error tolerance. For categorical attributes, we propose a baseline-normalized confidence score that measures how much more confident the attacker is about the true class than would be expected from class prevalence alone, and we summarize risk as the fraction of records exceeding a policy-defined threshold. This construction yields an interpretable, bounded risk metric that is robust to class imbalance, independent of any specific synthesizer, and applicable with arbitrary learning algorithms. We illustrate threshold calibration, uncertainty quantification, and comparative evaluation of synthetic data generators using simulations and real data. Our results show that RAPID provides a practical, attacker-realistic upper bound on attribute-inference disclosure risk that complements existing utility diagnostics and disclosure control frameworks.
\end{abstract}



\section{Introduction}
\label{sec:introduction}

Open research data (ORD) are increasingly recognized as essential for scientific transparency and reproducibility \citep{2015_Nosek}. Yet the proportion of shared datasets remains low: for instance, only 23\% of projects funded by the Swiss National Science Foundation currently provide ORD \citep{SNF_ORD2024}. Legal constraints and contractual usage restrictions often hinder the release of microdata, while technical barriers to anonymization remain substantial for many research teams.

Fully synthetic microdata \citep{Rubin93} -- datasets in which all records are simulated rather than perturbed copies of real individuals -- offer a promising solution to this problem. However, synthetic data also raise an immediate question for data stewards: how should disclosure risk be quantified when the primary threat is no longer re-identification, but an adversary's ability to infer sensitive attributes from the released data?

\subsection{The user perspective: data utility}

From the user's perspective, synthetic data must satisfy two core requirements: statistical similarity to the original data and structural plausibility. Beyond reproducing marginal distributions and associations, synthetic data must respect domain-specific constraints -- such as realistic household compositions, non-negative expenditures (e.g., on medication), internally consistent demographic characteristics (e.g., no underage individuals with adult children), and valid event sequences (e.g., a PhD obtained after a master's degree). Violations of such constraints can severely limit the credibility and usability of synthetic datasets, even when distributional similarity is high.

\begin{figure}[!htp]
    \centering
    \includegraphics[width=1\linewidth]{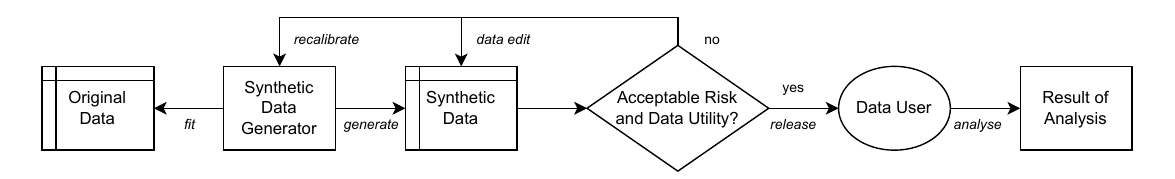}
    \caption{Synthetic data generation and evaluation workflow. Original data is used to train a synthetic data generator, which produces synthetic records. These records are evaluated for privacy risk and statistical utility. If acceptable, the synthetic data is released to data users for analysis; otherwise, the generator is recalibrated or the data is edited.}
    \label{fig:utility}
\end{figure}

Figure~\ref{fig:utility} illustrates a general workflow for generating synthetic data and assessing its fitness for use. Analysts typically aim to (i) answer predefined research questions by fitting statistical or machine learning models and (ii) explore the data to identify new patterns and relationships. These objectives are attainable only when the synthetic data approximate the original data not merely in distribution, but also in structure.

However, high analytical utility alone does not guarantee safe data release. From the data provider's perspective, increasing utility often entails preserving strong dependencies among variables -- precisely the information that may enable an adversary to infer sensitive attributes. Disclosure risk therefore does not necessarily decrease as synthetic data become more useful; in fact, it may increase. In practice, this tension is commonly conceptualized using Risk -- Utility (RU) maps \citep{Duncan01a} and more recent multivariate extensions \citep{Thees25pca}, which support informed decisions about acceptable trade-offs between analytical value and disclosure protection. Assessing inference disclosure risk in a way that is commensurate with modern, high-utility synthetic data remains a key methodological challenge.

\subsection{Overview of synthetic data generation methods}

Since \citet{Rubin93} introduced synthetic data as simulated draws from predictive distributions, a variety of generation methods have emerged. \emph{Conditional modeling} approaches synthesize variables sequentially, with each variable generated conditional on those already produced; implementations such as synthpop \citep{Nowok_2016_Synthpop} and simPop \citep{simPop} use tree-based methods, regression, or other learners. \emph{Joint modeling} approaches, including GANs and deep generative models, attempt to capture the full joint distribution simultaneously but require large datasets, careful tuning, and substantial computational resources \citep{mekonnen24,miletic24}, and may struggle with complex tabular structures \citep{thees24,ward25}.

Crucially, the same fidelity that makes synthetic data analytically useful also preserves the predictive relationships that enable attribute inference attacks. High-utility synthesis faithfully reproduces covariate -- outcome associations; an attacker can exploit these associations to infer sensitive attributes from quasi-identifiers. This tension between utility and privacy motivates the measure we propose.

While this paper focuses on synthetic data, the inferential disclosure risk framework applies equally to traditionally anonymized data or any released dataset: given the release, how accurately can an adversary infer sensitive attributes from quasi-identifiers?

\subsection{Disclosure risk measures}

Classical statistical disclosure control (SDC) distinguishes (i) identity disclosure (i.e., linking a record to a specific individual); (ii) attribute disclosure (i.e., correctly learning sensitive values); and
(iii) membership disclosure (i.e., learning whether an individual's data is part of a dataset) \citep{Hundepool_2012_Statistical}.

For fully synthetic microdata, identity disclosure risk is typically low \citep{templ14risk,emam20}, as synthetic records are not literal representations of individuals. As a result, most attention shifts toward analytical validity and attribute disclosure risk.

\paragraph{Existing attribute-disclosure diagnostics.}
Existing attribute-disclosure diagnostics fall into two broad categories:

\begin{itemize}
    \item Match-based measures, which evaluate how often synthetic values exactly match original values \citep{taub18}. Examples include DiSCO \citep{Raab_2025_Practical}, which flags records as at-risk when a quasi-identifier combination in the original data appears in the synthetic data with the same sensitive value. Such approaches are transparent and integrate naturally with classical SDC concepts, but they may be sensitive to class imbalance and do not directly model the inference process.
    \item Model-based measures, which train a predictive model on synthetic data and assess its performance on the original data. \citet{hittmeir20} proposed comparing prediction accuracy against a baseline that represents the attacker's prior knowledge (e.g., marginal class frequencies), thereby quantifying the additional risk introduced by the synthetic release. This baseline-comparison principle -- measuring how much better an attacker can do with the released data than without it -- is central to the approach we develop in this paper.
\end{itemize}

\paragraph{Attacker scenario.}
For model-based measures, the threat model assumes an attacker who trains a predictive model on the released synthetic data and applies it to individuals whose quasi-identifiers are known. Figure~\ref{fig:prediction} illustrates this scenario. If a model trained solely on synthetic data can reliably predict confidential attributes in the real data, the synthetic release carries privacy risk \citep{taub18, barrientos18, kwatra24}.

\paragraph{Inferential disclosure.}
The risk measure proposed here focuses on inferential disclosure risk -- also referred to as predictive disclosure \citep{Willenborg_2001_Elements} -- which occurs when the publication of microdata enables more accurate or more confident inferences about sensitive attributes than would have been possible without the release \citep{Duncan_1989_Risk, Hundepool_2012_Statistical}. This concept builds on Dalenius's foundational definition:
\begin{quote}
``If the release of the statistic $S$ makes it possible to determine the value [of a sensitive attribute] more accurately than is possible without access to $S$, a disclosure has taken place.'' \citep[p.~432]{Dalenius_1977_Methodology}
\end{quote}

Importantly, inferential disclosure can affect individuals who were not part of the original dataset and may concern information of which the affected individual is not even aware. For example, if the released data reveal a strong association between lifestyle indicators and disease risk, an attacker could infer elevated risk for a similar individual outside the dataset.

Although eliminating inferential disclosure entirely would require destroying all meaningful relationships between sensitive and non-sensitive variables \citep{Dwork_2010_Difficulties}, we argue that inferential risk warrants increased attention in the era of big data and artificial intelligence. \citet{Muhlhoff_2021_Predictive} notes that predictive privacy is violated when sensitive information is statistically estimated against an individual's will, provided that these predictions lead to differential treatment affecting their wellbeing or freedom.

\paragraph{The need for record-level assessment.}
While existing model-based measures typically provide aggregate, dataset-level risk summaries, we argue that record-level assessment is essential for targeted risk mitigation. Identifying which specific individuals face elevated disclosure risk enables data custodians to apply selective protections rather than uniform measures that may unnecessarily degrade utility. This motivates our proposed measure, RAPID, which we introduce after discussing the role of differential privacy.

\subsection{Differential privacy and attribute inference}

A natural question is whether differential privacy (DP) already addresses the disclosure risks we aim to measure. We argue that DP and RAPID address complementary concerns.

Differential privacy \citep{Dwork_2006_Calibrating} provides a formal guarantee of output stability: the probability of any particular output changes by at most a multiplicative factor $\exp(\varepsilon)$ when a single individual's data is added to or removed from the dataset \citep{domingo21}. This guarantee is independent of the intruder's background knowledge and does not rely on assumptions about data distributions or attacker capabilities.

Crucially, DP does not claim that an attacker cannot learn sensitive information about an individual. Rather, it guarantees that the attacker's information gain is essentially the same whether or not a specific individual's data are included in the dataset \citep{MuralidharRuggles2024}. The rationale is that if inclusion does not matter, then no individual-specific privacy breach can occur. This makes DP particularly effective at preventing \emph{membership inference attacks}, where the adversary tries to determine whether a specific individual was present in the training data.

However, \emph{attribute inference attacks} exploit a different mechanism. Rather than asking ``Was this person in the dataset?'', the attacker asks ``What is this person's sensitive attribute, given the released data and what I know about them?'' DP does not directly limit the accuracy of such inferences, because the risk stems from population-level patterns that the data preserve, not from any individual's participation. If strong correlations exist in the population -- e.g., between age, education, and disease status -- then data released under DP (or any other anonymization method) may still encode those relationships, enabling accurate attribute inference \citep{BlancoJusticia2022,muralidhar23}.

This limitation is not unique to DP. Traditional anonymization methods face the same fundamental tension: preserving analytical utility requires preserving statistical relationships, but those same relationships enable inferential attacks. The question of whether releasing data increases an attacker's ability to infer sensitive attributes -- relative to what could be inferred from background knowledge alone -- applies regardless of the anonymization technique used. Moreover, DP faces a practical dilemma: stringent privacy guarantees (small $\varepsilon$) require noise levels that can render outputs analytically useless, while relaxed guarantees (large $\varepsilon$) offer little meaningful protection \citep{domingo21}.

For this reason, DP (and anonymization more broadly) should be complemented with explicit, scenario-based disclosure risk assessments that directly measure attribute-inference vulnerability. As \citet{DomingoFerrer_2025_Statistical} argue, empirical disclosure risk assessment remains as unavoidable for synthetic or DP-protected data as it was under traditional utility-first approaches. RAPID addresses this need by quantifying how accurately an attacker could infer sensitive attributes from the released data, providing a practical complement to formal privacy guarantees.

\paragraph{Relation to membership disclosure and DP baselines.}

Recent work on \emph{membership disclosure} situates attribute-inference risk within a differential privacy framework by comparing two anonymized datasets that differ only in the presence of a single individual -- a ``member'' and a ``non-member'' version -- and measuring the marginal improvement in inference accuracy when the individual is included \citep[e.g.,][]{francis2025betterattributeinferencevulnerability}. These approaches aim to quantify per-person \emph{privacy loss due to inclusion}, grounded in DP's stability guarantee.

RAPID addresses a different question. While membership-based evaluations focus on individual-level stability (``Does my inclusion change the risk?''), RAPID focuses on population-level vulnerability (``How often could an attacker be confidently correct?''). Both approaches contribute to understanding disclosure risk, but they serve distinct purposes: membership-focused methods are suited to certifying DP-style guarantees, while RAPID supports practical risk--utility assessment for public-use data releases.

\paragraph{What RAPID is and is not.}
{To avoid misinterpretation, we state explicitly what RAPID provides and what it does not claim. RAPID \emph{is} a scenario-based, attacker-realistic, empirical diagnostic that quantifies realized inferential vulnerability under a specified predictive model. It measures how often an attacker could confidently infer sensitive attributes from released data, conditional on the predictive structure those data preserve. RAPID \emph{is not} a formal privacy guarantee, a differential privacy analogue, or a worst-case bound. Rather, RAPID complements formal guarantees by offering transparent, actionable risk assessment that supports informed release decisions -- answering the practical question of whether released data enable inference attacks that data custodians would consider unacceptable.}

\subsection{Contributions}

This paper makes the following contributions.

\begin{enumerate}
\item \textbf{A new disclosure risk measure under a realistic threat model.}
We propose RAPID (Risk of Attribute Prediction--Induced Disclosure), a measure of attribute-inference disclosure risk for anonymized microdata, with a focus on fully synthetic data. RAPID quantifies the proportion of records for which an attacker can confidently infer sensitive attributes. Unlike differential privacy, which provides a worst-case bound on how much an individual's inclusion can influence the released data, RAPID measures how often inference succeeds across the released dataset -- a practical summary of disclosure vulnerability that data custodians can directly interpret and compare.

Our threat model assumes an intruder who has access only to the released data and to quasi-identifiers of target individuals, but no auxiliary sample of real data for validation or calibration. This reflects the conditions faced by external analysts of public-use files and avoids the practical limitations of holdout-based evaluations, whose conclusions depend heavily on holdout size and representativeness \citep{hittmeir20,PlatzerReutterer2021Holdout}. When evaluating risk, the data custodian adopts a conservative worst-case assumption: that the attacker knows the quasi-identifiers of all individuals in the original dataset (but not their sensitive attributes). RAPID then quantifies how often the attacker could successfully infer those attributes using only the released synthetic data. This worst-case framing means that if RAPID indicates acceptable risk, the risk is acceptable for any realistic attacker who knows fewer individuals. Whether these risk estimates generalize beyond the original sample to a broader population is a question of representativeness -- a consideration common to all empirical analyses, not a limitation specific to RAPID. RAPID directly operationalizes the attacker's objective: confidently inferring sensitive attributes of real individuals using models trained solely on the released data. This prediction-based vulnerability corresponds to \emph{inferential disclosure} in the classical SDC taxonomy \citep{Duncan_1989_Risk,Hundepool_2012_Statistical}. (See Section~\ref{sec:rapid}.)

\item \textbf{Baseline-normalized confidence scoring for categorical attributes.}
We introduce a normalized gain measure that compares an attacker's predicted probability for the true class to a baseline determined by class prevalence in the original data (cf. \citeauthor{Slokom_2022_When}, \citeyear{Slokom_2022_When}). This yields a calibrated notion of \emph{confidence beyond chance} that explicitly accounts for class imbalance and avoids overstating risk in skewed distributions. By focusing on confidence-adjusted correctness rather than raw accuracy, RAPID captures the extent to which released data enable \emph{meaningful} inference about sensitive attributes. (See Section~\ref{sec:rapid-cat} and Section~\ref{sec:toy}.)

\item \textbf{A unified framework for categorical and continuous sensitive attributes.}
RAPID provides a consistent formulation for both discrete and continuous confidential variables. For categorical attributes, inference risk is quantified via baseline-normalized prediction confidence; for continuous attributes, it is defined through tolerance-based relative prediction error. This unified framework allows inference risk to be assessed consistently across variable types without discretizing continuous outcomes, introducing arbitrary binning, or requiring counterfactual datasets that compare member versus non-member scenarios. (See Sections~\ref{sec:rapid-cat}--\ref{sec:rapid-cont} and Section~\ref{sec:toy}.)

\item \textbf{Threshold-based, policy-interpretable risk summaries.}
By summarizing disclosure risk as the proportion of records exceeding a confidence or accuracy threshold, RAPID yields interpretable, bounded metrics that are easy to communicate and tunable to institutional or regulatory risk tolerances. This formulation aligns naturally with risk--utility decision frameworks commonly used in statistical disclosure control and supports consistent comparison across different synthesizers, parameter settings, and data releases. (See Algorithm~\ref{alg:rapid} and Section~\ref{sec:properties}.)

\item \textbf{Record-level risk indicators enabling diagnostic analysis.}
Although RAPID reports an aggregate disclosure metric, it is constructed from per-record risk indicators. This design enables granular analyses of inference vulnerability, including the identification of high-risk records, subgroup-specific risk patterns, and combinations of quasi-identifiers that disproportionately contribute to disclosure risk. These record-level signals help data curators understand the drivers of risk and apply targeted mitigation strategies. (See Section~\ref{qi-attribution} and Figure~\ref{fig:sim3}.)

\item \textbf{Empirical validation and robustness analysis.}
Through simulation studies and real-data illustrations, we demonstrate how RAPID scales with dependency strength between quasi-identifiers and sensitive attributes, how threshold choice affects risk classification, and how results remain stable across a range of attacker models. We also show how RAPID can be combined with holdout-based assessments when auxiliary real data are available, supporting internal benchmarking without altering the core threat model. (See Sections~\ref{sec:adult} and~\ref{sec:simulations}.)

\item  \textbf{Open-source implementation.}
RAPID is implemented in R and integrates
with existing synthetic data generation pipelines.
The implementation supports multiple attacker models, bootstrap confidence intervals,
and diagnostic outputs. By focusing on population-level vulnerability rather
than per-individual privacy loss, RAPID complements formal privacy frameworks
such as differential privacy and is particularly suited to evaluating public-use
microdata releases. (See Section~\ref{subsec:software}).
\end{enumerate}

\subsection{Motivating example}
\label{sec:motivation-attack}

To illustrate the inference threat RAPID addresses, we conducted a simple attack experiment. Using the \texttt{eusilc} public-use file -- a close-to-reality synthetic population based on the European Union Statistics on Income and Living Conditions (EU-SILC), which contains complex household structures and a rich set of demographic and socioeconomic variables \citep{Alfons_2011_Simulation} -- we generated a fully synthetic dataset with \texttt{synthpop} \citep{Nowok_2016_Synthpop}. We then trained random forest models \citep{Wright_2017_Ranger} on the synthetic data to predict two sensitive attributes: the categorical variable \texttt{marital\_status} and the continuous variable \texttt{income}.

Applying these synthetic-trained models to real covariate values reveals meaningful disclosure risk. For \texttt{marital\_status}, the classifier achieved 82\% accuracy on real data despite never seeing true labels during training. For \texttt{income}, many predictions fell close to true values, indicating that an attacker could approximate real incomes within a practically relevant range.

This experiment demonstrates that standard metrics like accuracy or mean squared error, while useful for evaluating predictive performance, do not adequately characterize disclosure risk. They ignore the attacker's \emph{confidence} in correct predictions, are sensitive to class imbalance, and provide no policy-interpretable threshold. These limitations motivate the formal development of RAPID in the following section, which provides a unified framework for both categorical and continuous sensitive attributes.


\section{The RAPID measure}
\label{sec:rapid}

This section formalizes RAPID (Risk of Attribute Prediction--Induced Disclosure). Recall that our threat model assumes an attacker who trains a predictive model on the released data and applies it to individuals whose quasi-identifiers are known (cf.\ Figure~\ref{fig:prediction}). RAPID quantifies how often such an attacker can make \emph{confidently correct} inferences about a sensitive attribute.

The key idea is to compare the attacker's prediction performance against a baseline that ignores quasi-identifiers entirely.
For categorical attributes, this baseline is the marginal frequency of each class in the original data -- the probability an attacker would assign to the
correct class by simply knowing class prevalences without using any quasi-identifier information. For continuous attributes, the baseline is a
reference prediction error.

A record is flagged as at-risk only when the attacker's performance substantially exceeds this baseline, ensuring that RAPID captures genuine
information leakage rather than artifacts of class imbalance or distributional properties.

\subsection{Setup and notation}

Let the original microdata be denoted by
\[
\mathbf{Z} = [\mathbf{X}_Q,\; \mathbf{X}_U,\; \mathbf{y}] \quad ,
\]
where $\mathbf{X}_Q \in \mathbb{R}^{n \times p_Q}$ represents quasi-identifiers known to an attacker, $\mathbf{X}_U \in \mathbb{R}^{n \times p_U}$ denotes additional attributes not available to the attacker, and $\mathbf{y} \in \mathbb{R}^n$ is the confidential sensitive attribute.

The released (e.g., synthetic) dataset is
\[
\mathbf{Z}^{(s)} = [\mathbf{X}_Q^{(s)},\; \mathbf{X}_U^{(s)},\; \mathbf{y}^{(s)}] \quad .
\]
An attacker observes the released data $\mathbf{Z}^{(s)}$ together with quasi-identifiers $\mathbf{X}_Q$ of target individuals (e.g., from external sources), but has no access to $\mathbf{X}_U$ or $\mathbf{y}$.

The attacker trains a predictive model $\mathcal{M}$ on $(\mathbf{X}_Q^{(s)}, \mathbf{y}^{(s)})$ to obtain parameter estimates $\hat{\Theta}^{(s)}$, then applies this model to $\mathbf{X}_Q$ to produce predictions $\hat{\mathbf{y}}$ of the sensitive attribute.

For notational simplicity, we write $\mathbf{Z} = [\mathbf{X},\; \mathbf{y}]$, where $\mathbf{X}$ denotes quasi-identifiers, since $\mathbf{X}_U$ plays no role in risk estimation. By default, we evaluate risk across all $n$ records in the original data. However, RAPID's record-level design allows risk assessment for any target set -- a specific subpopulation, a sample, or even individuals not in the original data whose quasi-identifiers are known.

\begin{figure}[!htp]
    \centering
    \includegraphics[width=0.7\linewidth]{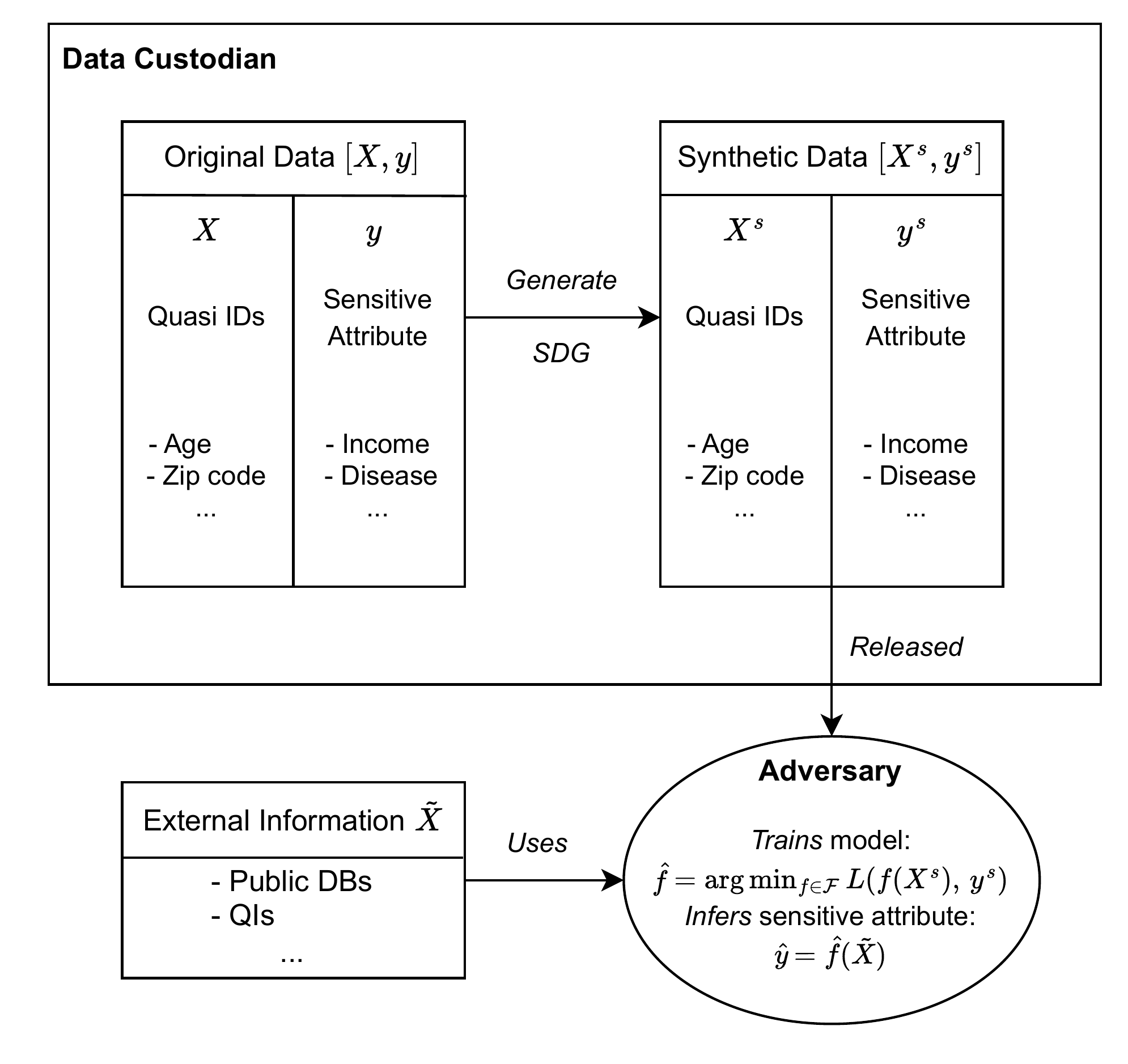}
\caption{Inferential disclosure threat model. A data custodian releases synthetic data [$\mathbf{X}^s$, $\mathbf{y}^s$] generated from original data containing quasi-identifiers ($\mathbf{X}$) and a sensitive attribute ($\mathbf{y}$). An adversary with access to the released synthetic data and external knowledge of individuals' quasi-identifiers ($\tilde{\mathbf{X}}$, which may or may not include individuals from the original sample) trains a predictive model to infer sensitive attributes.}
    \label{fig:prediction}
\end{figure}

Figure~\ref{fig:prediction} illustrates this threat model from a data custodian's viewpoint.

\subsection{RAPID for a categorical sensitive attribute}
\label{sec:rapid-cat}

When $\mathbf{y}$ is categorical, we define risk in terms of the model-assigned probability to the true class label. For each record $i$, the attacker's model assigns probability
\[
g_i = \Pr(\hat{y}_i = y_i \mid \mathbf{x}_i, \hat{\Theta}^{(s)})
\]
to the true class $y_i$. To evaluate whether this confidence is unusually high, we compare it against a baseline $b_i$ defined as the marginal proportion of class $y_i$ in the original data:
\[
b_i = \frac{1}{n}\sum_{j=1}^{n} \mathbb{I}(y_j = y_i)
\]
This baseline represents the prediction confidence achievable by simply guessing according to the marginal class distribution, without using any quasi-identifier information. We use the original marginals (rather than synthetic marginals) to ensure a meaningful, consistent baseline: RAPID measures improvement over the true population rates, not over potentially distorted synthetic marginals. This makes risk estimates comparable across different synthesizers and avoids conflating poor marginal preservation with low disclosure risk.

We compute a normalized gain score,
\[
r_i = \frac{g_i - b_i}{1 - b_i}
\]
that measures the improvement in prediction confidence over baseline, normalized by the maximum possible improvement (reaching perfect confidence $g_i = 1$).
The score satisfies:
\begin{itemize}
    \item $r_i < 0$: model performs worse than baseline
    \item $r_i = 0$: model performs at baseline (no information gain)
    \item $0 < r_i < 1$: partial information gain from quasi-identifiers
    \item $r_i = 1$: perfect prediction (complete disclosure)
\end{itemize}

A record is considered at risk if $r_i > \tau$, where $\tau \in (0,1)$ is a policy-defined threshold. We recommend $\tau = 0.3$ as a default, meaning a record is flagged when the attacker achieves at least 30\% of the maximum possible improvement over baseline.
This threshold balances sensitivity (detecting meaningful inference gains)
with specificity (avoiding false positives from minor improvements);
Section~\ref{sec:simulations} provides empirical guidance on threshold selection.

The categorical RAPID metric is then:
\[
\text{RAPID}^{\text{cat}}(\tau) = \frac{1}{n}\sum_{i=1}^{n} \mathbb{I}(r_i > \tau)
\]
representing the proportion of records for which the released data enable inference substantially better than the baseline rate.

\subsection{RAPID for a continuous sensitive attribute}
\label{sec:rapid-cont}

When the confidential attribute $\mathbf{y}$ is continuous, we assess whether the model prediction $\hat{y}_i$ is sufficiently close to the true value $y_i$. This is operationalised via a prediction error $e_i$, which is compared to a fixed global threshold $\varepsilon$.
A record $i$ is considered \textit{at risk} if $\text{at-risk}_i = \mathbb{I}(e_i < \varepsilon)$, where $\varepsilon$ is a fixed tolerance level reflecting how close a prediction must be to the true value to constitute a disclosure risk.
Several error metrics can be used, depending on the application context. Table~\ref{tab:error-metrics-continuous} summarises suitable choices. Importantly, all proposed error measures are dimensionless and comparable, such that a single threshold $\varepsilon$ can be applied consistently across all records and variables.

\begin{table}[!htbp]
\centering
\caption{Error metrics for continuous sensitive attributes. Records are flagged as at risk if $e_i < \varepsilon$, where $\varepsilon$ is a user-specified threshold. The smoothing constant $\delta > 0$ prevents division by zero (default: $\delta = 0.01$, or scaled to the data range).}
\label{tab:error-metrics-continuous}
\begin{tabular}{p{5.5cm} p{9.5cm}}
\toprule
\textbf{Error metric $e_i$} &
\textbf{Description, usage, and examples} \\
\midrule

\textbf{Symmetric relative error} (default)
\[
\frac{2|y_i - \hat{y}_i|}{|y_i| + |\hat{y}_i| + 2\delta}
\]
&
Robust to scale, accounts for both true and predicted values in error calculation.

\textit{Example:} For income data with $\varepsilon = 0.05$ (5\%), a record with true income \$50,000 and predicted income \$52,000 has error $\approx 3.9\%$ and is flagged as at risk.

\textit{Use when:} Default choice for percentage-based error measurement. \\

\midrule

\textbf{Stabilised relative error}
\[
\frac{|y_i - \hat{y}_i|}{|y_i| + \delta}
\]
&
Simpler formula than symmetric. Produces similar results for most values but larger errors for very small true values.

\textit{Example:} For income data with $\varepsilon = 0.10$ (10\%), a record with true income \$500 and predicted income \$450 has error $\approx 10\%$ and is at risk.

\textit{Use when:} Traditional in some fields or when simpler formula is preferred. \\

\midrule

\textbf{Absolute error}
\[
|y_i - \hat{y}_i|
\]
&
No percentage interpretation; errors measured in original units of $y$. Threshold $\varepsilon$ must be on the same scale.

\textit{Example:} For age data with $\varepsilon = 2$ years, a record with true age 65 and predicted age 66 has error 1 year and is flagged as at risk.

\textit{Use when:} The meaningful threshold is "within X units" rather than "within X\%" (e.g., age within 2 years, temperature within 1°C, dates within 7 days). \\

\bottomrule
\end{tabular}
\end{table}

In all cases, the continuous RAPID metric is:
\[
\text{RAPID}^{\text{cont}}(\varepsilon) = \frac{1}{n}\sum_{i=1}^{n} \mathbb{I}(e_i < \varepsilon)
\]
representing the proportion of records where predictions fall within the specified error tolerance.
Note that while aggregate error metrics (MAE, RMSE) are insufficient for disclosure assessment -- as they do not identify which records are at risk -- they can inform the choice of $\varepsilon$ by characterizing typical prediction accuracy.

\subsection{Computation}
\label{sec:computation}

Algorithm~\ref{alg:rapid} formalizes the RAPID evaluation protocol: train a predictive model on the released data, apply it to target quasi-identifiers, compute per-record risk scores, and aggregate to obtain RAPID.

\begin{algorithm}[H]
\caption{RAPID Evaluation Protocol}
\label{alg:rapid}
\begin{algorithmic}[1]
\Require Original data $\mathbf{Z} = [\mathbf{X}, \mathbf{y}]$; released data $\mathbf{Z}^{(s)}$
\Require Thresholds $\tau \in (0,1)$ (categorical; default $\tau = 0.3$) and $\varepsilon > 0$ (continuous; default $\varepsilon = 10\%$)

\State Fit predictive model $\mathcal{M}$ on $\mathbf{Z}^{(s)}$ to obtain $\hat{\Theta}^{(s)}$

\State Compute predictions $\hat{\mathbf{y}} = \mathcal{M}(\mathbf{X}; \hat{\Theta}^{(s)})$

\For{$i = 1, \ldots, n$}
    \If{$y$ is categorical}
        \State $g_i \gets \Pr(\hat{y}_i = y_i \mid \mathbf{x}_i)$
        \State $b_i \gets n^{-1} \sum_{j=1}^n \mathbb{I}(y_j = y_i)$
        \State $r_i \gets (g_i - b_i)/(1 - b_i)$
        \State $I_i \gets \mathbb{I}(r_i > \tau)$
    \Else
        \State $e_i \gets 2|y_i - \hat{y}_i|/(|y_i| + |\hat{y}_i| + 2\delta)$
        \State $I_i \gets \mathbb{I}(e_i < \varepsilon)$
    \EndIf
\EndFor

\State \textbf{Output:}
record-level risks $\{r_i\}$ or $\{e_i\}$,
indicators $\{I_i\}$, and
\[
\text{RAPID} \gets \frac{1}{n} \sum_{i=1}^n I_i
\]
\end{algorithmic}
\end{algorithm}

RAPID is thus an empirical estimate of the probability that a randomly selected record is subject to successful attribute inference under the specified attacker model.

When evaluating multiple models $\mathcal{M} \in \mathcal{S}$, report both
the average and conservative envelope:
\[
\overline{\text{RAPID}}(\cdot) = \frac{1}{|\mathcal{S}|}\sum_{m\in \mathcal{S}}\text{RAPID}^{(m)}(\cdot),
\qquad
\text{RAPID}_{\max}(\cdot) = \max_{m\in \mathcal{S}}\text{RAPID}^{(m)}(\cdot)
\]

\subsection{Implementation considerations}
\label{sec:implementation}

RAPID is applicable to any algorithm producing class probabilities or point predictions. To reflect a strong attacker, we recommend
evaluating powerful learners (e.g., random forests, gradient boosting) and reporting the maximum observed risk. When multiple released datasets are
available, risk can be averaged across them or reported as the worst-case value, depending on desired conservativeness. Thresholds $\tau$ and $\varepsilon$
should reflect policy requirements; we recommend defaults of $\tau = 0.3$ and $\varepsilon = 10\%$, which we validate empirically in Section~\ref{sec:simulations}. For data-driven selection, one can permute $\mathbf{y}$, recompute RAPID, and choose the threshold placing observed risk above the 95th percentile of the
resulting null distribution. Uncertainty can be quantified via bootstrap resampling or, treating RAPID as a binomial proportion, using Wilson score \citep{wilson1927probable} or Clopper--Pearson confidence intervals \citep{clopper1934use}.

\paragraph{Evaluating synthetic data generators.}
\label{sec:cv}

To assess the general disclosure risk of a synthetic data generator (SDG), we recommend $k$-fold cross-validation: partition the original data into $k$ folds (stratified by $\mathbf{y}$ if categorical), generate synthetic data from $k$-$1$ folds, and evaluate RAPID on the held-out fold. Aggregating across folds yields the expected risk $\mathbb{E}$[RAPID] with confidence intervals, providing a robust assessment of the SDG's average disclosure risk. This approach is implemented via the \texttt{rapid\_synthesizer\_cv()} function and is particularly useful for selecting between alternative synthesis methods (e.g., CART vs.\ parametric) or optimizing hyperparameters. Once a method is selected, standard RAPID should be applied to the final synthetic data product to obtain record-level risk assessments prior to the data release.

\paragraph{Computational complexity.}
The computational cost of RAPID is dominated by the model training step (line~1 in Algorithm~\ref{alg:rapid}), with subsequent prediction and risk evaluation requiring only $O(n)$ operations. For random forests, training complexity is $O(n \cdot p \cdot T \cdot \log n)$ where $n$ is the number of records, $p$ the number of quasi-identifiers, and $T$ the number of trees; prediction scales as $O(n \cdot T \cdot d)$ where $d$ is tree depth. Gradient boosting machines exhibit similar $O(n \cdot p \cdot T)$ training complexity. Linear models and logistic regression require $O(n \cdot p^2)$ operations via the normal equations or iterative fitting, while CART trees scale as $O(n \cdot p \cdot \log n)$.

In practice, RAPID evaluation completes within seconds for datasets of 10,000 records and remains tractable for $n > 100{,}000$ with default random forest settings ($T = 500$ trees). Memory usage is dominated by the fitted model object; random forests with many trees on high-dimensional data may require several hundred megabytes. When evaluating multiple attacker models or bootstrap replicates, computations are embarrassingly parallel across models, enabling efficient use of multi-core systems. 

\subsection{Properties}
\label{sec:properties}

RAPID has several structural properties that follow directly from its definition.
First, it targets successful and confident attribute inference events rather than aggregate predictive accuracy, aligning the risk measure with inferential disclosure in the classical SDC taxonomy. Second, the normalization by the empirical base rate ensures that categorical risk scores are invariant to class imbalance, so that common outcomes do not spuriously inflate disclosure risk.

Third, RAPID is bounded in $[0,1]$, as it is defined as the empirical mean of record-level disclosure indicators. This facilitates comparison across datasets, synthesizers, and attacker models. Fourth, the construction is flexible with respect to the attacker model: any predictive model capable of producing point predictions or class probabilities may be used, allowing RAPID to be evaluated under strong and adaptive attacker assumptions.

Finally, RAPID is monotone in the decision thresholds $\tau$ and $\varepsilon$, respectively. Increasing the categorical threshold $\tau$ or decreasing the continuous tolerance $\varepsilon$ can only reduce the number of records classified as at risk. This monotonicity supports transparent sensitivity analysis and makes RAPID well suited for practical disclosure control.

Because RAPID operates at the record level, it also enables subgroup-specific and conditional risk analysis by restricting the indicator average to subsets of the quasi-identifier space (see, e.g., Figure~\ref{fig:sim3}).

\section{Toy example}
\label{sec:toy}

We demonstrate the categorical RAPID metric on a simple example. Consider a dataset with 100 records where class \texttt{healthy} has 60\% prevalence
(marginal proportion). A model trained on released data is applied to three original records, all truly belonging to class \texttt{healthy}:

\begin{itemize}
    \item Record 1: $g_1 = \Pr(\hat{y}_1 = \text{healthy} \mid \mathbf{x}_1) = 0.70$
    \item Record 2: $g_2 = \Pr(\hat{y}_2 = \text{healthy} \mid \mathbf{x}_2) = 0.85$
    \item Record 3: $g_3 = \Pr(\hat{y}_3 = \text{healthy} \mid \mathbf{x}_3) = 0.55$
\end{itemize}

The baseline is the marginal frequency of class \texttt{healthy} in the original data:
\[
b_i = 0.60 \quad \text{for all three records (same class)}
\]

We compute the normalized gain for each record:
\begin{align*}
r_1 &= \frac{0.70 - 0.60}{1 - 0.60} = \frac{0.10}{0.40} = 0.25 \\
r_2 &= \frac{0.85 - 0.60}{1 - 0.60} = \frac{0.25}{0.40} = 0.625 \\
r_3 &= \frac{0.55 - 0.60}{1 - 0.60} = \frac{-0.05}{0.40} = -0.125
\end{align*}

\textbf{Interpretation:}
\begin{itemize}
    \item Record 1: 25\% of maximum improvement over baseline
    \item Record 2: 62.5\% of maximum improvement (high confidence!)
    \item Record 3: Below baseline (model worse than guessing)
\end{itemize}

Using the default threshold $\tau = 0.3$, we flag records where $r_i > 0.3$:
\[
\mathbb{I}\{r_1 > 0.3\} = 0, \quad
\mathbb{I}\{r_2 > 0.3\} = 1, \quad
\mathbb{I}\{r_3 > 0.3\} = 0
\]

The RAPID metric is:
\[
\text{RAPID}^{\text{cat}}(0.3) = \frac{1}{3}(0 + 1 + 0) = 0.33
\]

This indicates that 33\% of records have predictions substantially better than the baseline rate. Record 2, with 62.5\% normalized gain, represents a high
disclosure risk: the released data enables confident inference beyond what marginal class frequencies alone would allow. This example illustrates how
RAPID identifies records where quasi-identifiers provide meaningful information gain, rather than merely measuring prediction accuracy.

\vspace{1em}
\noindent
\textbf{Example (continuous attribute):}

Now consider the case where the sensitive variable is continuous, such as income. Suppose an attacker trains a regression model on released data and applies it
to the original covariates of three individuals. Let the true incomes and the model's predictions be:

\begin{itemize}
    \item Record 1: $y_1 = 50{,}000$, $\hat{y}_1 = 47{,}000$
    \item Record 2: $y_2 = 35{,}000$, $\hat{y}_2 = 39{,}000$
    \item Record 3: $y_3 = 80{,}000$, $\hat{y}_3 = 90{,}000$
\end{itemize}

The stabilised relative prediction errors (as percentages of true values (assuming $\delta = 0$) are:
\begin{align*}
e_1 &= \frac{|50{,}000 - 47{,}000|}{50{,}000} \times 100 = 6\% \\
e_2 &= \frac{|35{,}000 - 39{,}000|}{35{,}000} \times 100 = 11.4\% \\
e_3 &= \frac{|80{,}000 - 90{,}000|}{80{,}000} \times 100 = 12.5\%
\end{align*}
Using a threshold of $\varepsilon = 10\%$, we flag records where predictions
fall within 10\% of the true value:
\[
\mathbb{I}\{e_1 < 10\%\} = 1, \quad
\mathbb{I}\{e_2 < 10\%\} = 0, \quad
\mathbb{I}\{e_3 < 10\%\} = 0
\]

The continuous RAPID metric is:
\[
\text{RAPID}^{\text{cont}}(10\%) = \frac{1}{3}(1 + 0 + 0) = 0.33
\]

This indicates that for one-third of individuals, the attacker's model predicted income within 10\% relative error. Record 1, with only 6\% error, represents a disclosure risk: the released data enables an attacker to infer income with high precision. This form of attribute disclosure is directly relevant for risk assessment but would not be captured by aggregate metrics like mean absolute error alone.

\subsection{Software and defaults}
\label{subsec:software}

The proposed risk measure is implemented in the \textsf{R} package \texttt{RAPID}, publicly available on GitHub \citep{RAPID}. The package builds on well-established libraries: \texttt{ranger} \citep{Wright_2017_Ranger} for random forests, which provides native parallelization and memory-efficient storage, \texttt{glm} for logistic regression, and \texttt{xgboost} \citep{Chen_2016_XGBoost} for gradient boosting. It provides functions for computing RAPID for both categorical and continuous sensitive attributes, including baseline normalization, threshold calibration, and bootstrap-based uncertainty quantification.

Unless otherwise noted, we adopt the following default settings in our experiments: the attacker model $\mathcal{M}$ is a random forest with 500 trees and probabilistic outputs enabled. For categorical attributes, we use the default threshold $\tau = 0.3$; for continuous attributes, we use the default relative error tolerance $\varepsilon = 0.10$. Uncertainty is quantified via a nonparametric bootstrap over the original dataset (500 replicates), with percentile-based confidence intervals.

\section{Real data illustration: UCI Adult (Census Income)}
\label{sec:adult}

We illustrate the workflow on the training portion of the \emph{Adult} dataset (UCI Machine Learning Repository; 32{,}561 rows, 15 attributes; binary confidential attribute $\mathbf{y} = \mathbb{I}(\text{income} > \$50\text{K})$) \citep{adult_2}. The covariates, $\mathbf{X}$, include age, education, hours-per-week, marital status, etc. We treat $\mathbf{y}$ as the confidential, sensitive variable and $\mathbf{X}$ as potentially known quasi-identifiers.

\paragraph{Pre-processing.}
We removed the census sampling weight (\texttt{fnlwgt}) and the redundant categorical education variable (retaining \texttt{education.num}) and converted character variables to factors.

\paragraph{Synthesizers.}
We consider a CART-based tabular synthesizer (as implemented in \texttt{synthpop} \citep{Nowok_2016_Synthpop}), which is trained solely on $\mathbf{Z}$ and produces $M=5$ synthetic replicates.

\paragraph{Attack models.}
We evaluate an attacker suite $\mathcal{S}=\{$RF, GBM, $\ell_1$-logistic$\}$, trained on each synthetic replicate and scored on the real covariates $\mathbf{X}$.

\begin{figure}[!htp]
\centering
\includegraphics[width=1\textwidth]{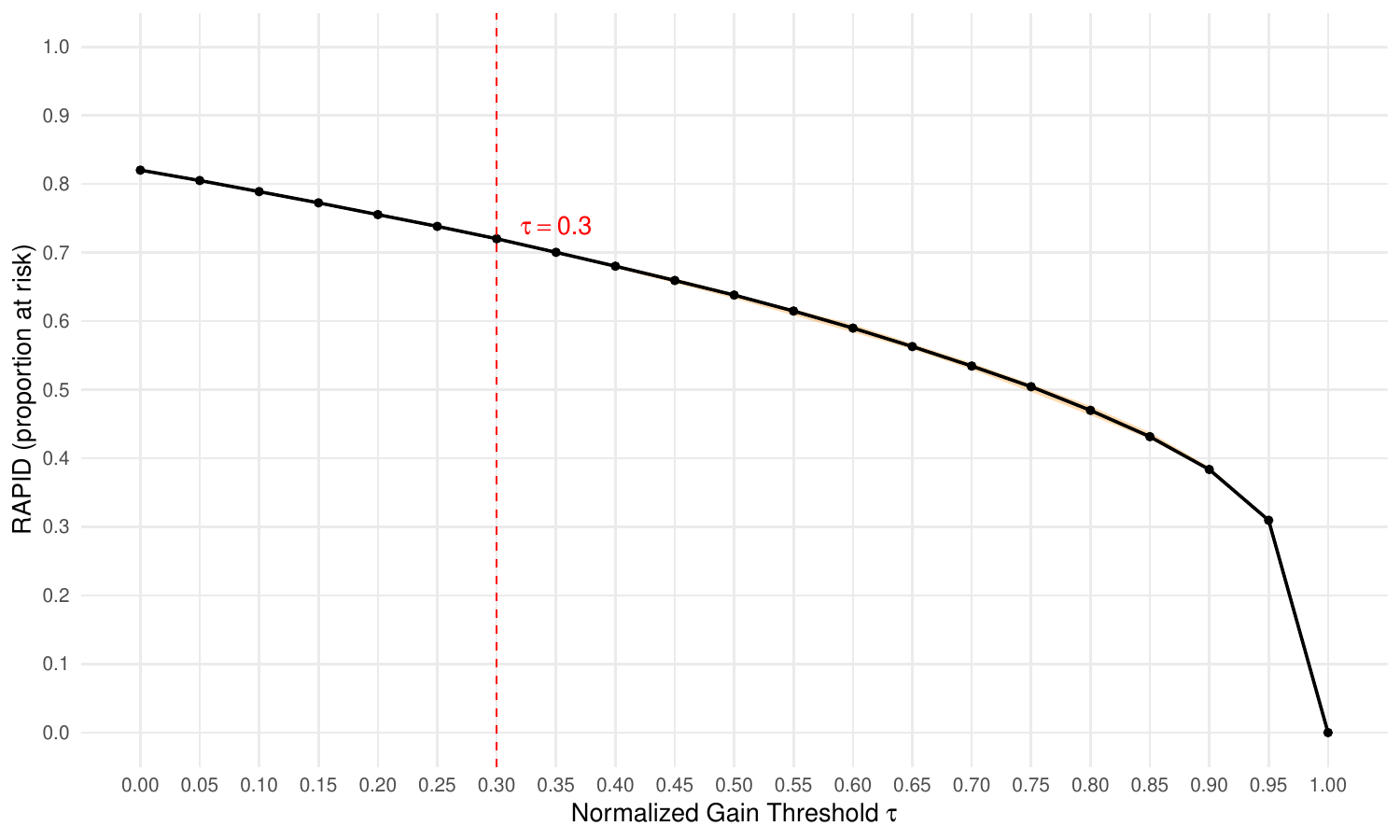}
\caption{Threshold sensitivity curve for the UCI Adult dataset.
RAPID (proportion of records at risk) as a function of the normalized gain threshold $\tau$,
averaged across 5 synthetic replicates generated via CART synthesis using synthpop \citep{raab2024synthpop}.
The vertical dashed line marks the default threshold ($\tau=0.3$).
The curve demonstrates how disclosure risk decreases as stricter thresholds are imposed,
with approximately 70\% of records flagged at $\tau=0.3$ and less than 5\% at $\tau=0.9$.
Shaded region indicates range across replicates.}
\label{fig:rcir-curve}
\end{figure}

\paragraph{Metrics and reporting.}
We compute $\mathrm{RAPID}^{\text{cat}}(\tau)$ across a range of threshold values and visualize the results as a threshold curve (Figure~\ref{fig:rcir-curve}). For each synthetic replicate, we average RAPID across the $M=5$ replicates and report 95\% bootstrap confidence intervals. We additionally stratify RAPID by true class $y$ to reveal whether disclosure risk differs across outcome categories -- for example, whether high-income individuals are more identifiable than low-income individuals.

\paragraph{Sensitivity and diagnostics.}
To understand how RAPID responds to threshold choices, we plot $\mathrm{RAPID}^{\text{cat}}(\tau)$ across a grid of $\tau$ values (Figure~\ref{fig:rcir-curve}), visualizing how disclosure risk decays as stricter normalized gain thresholds are imposed.

Practitioners may consider additional diagnostics:
\begin{itemize}
\item \textbf{Class balance:} Reporting the baseline probability $b_k$ for each class $k$ exposes the influence of class prevalence on the normalized gain.
\item \textbf{Joint utility--risk view:} To contextualize disclosure risks, utility metrics such as predictive accuracy of models trained on $\mathbf{Z}^{(s)}$ but evaluated on $\mathbf{Z}$ can be reported alongside RAPID.
\end{itemize}

To assess attribute inference risk, we applied RAPID to the training portion of the UCI Adult dataset using the binary variable \texttt{income} as the sensitive attribute. Five synthetic datasets were generated via the CART synthesizer (\texttt{synthpop} package). For each replicate, we trained a random forest attacker model on the synthetic data to infer the sensitive attribute from quasi-identifiers, then evaluated the model's predictions on the original dataset.

We computed RAPID across a range of threshold values $\tau \in [0, 1]$ to examine how disclosure risk varies with the stringency of the threshold.
To quantify uncertainty, we performed non-parametric bootstrapping with $R = 1000$
resamples from the original dataset, providing robust percentile-based confidence intervals.
The RAPID score was calculated based on random forest predictions,
reflecting a strong attacker model. 

Figure~\ref{fig:rcir-curve} shows how RAPID varies with the normalized gain threshold $\tau$. As $\tau$ increases from 0 to 1, fewer records exceed the threshold, demonstrating the monotonicity property discussed in Section~\ref{sec:properties}. This threshold curve enables the data provider
to calibrate disclosure risk according to their privacy requirements.

Table~\ref{tab:rapid-adult} reports RAPID values at the default threshold $\tau = 0.3$ for each of the five synthetic replicates. The narrow confidence intervals, obtained via non-parametric bootstrapping with $R = 500$ resamples, indicate stable risk estimates across replicates. Approximately 72\% of records exhibit normalized gains exceeding the threshold, suggesting substantial attribute-inference risk for this dataset under the CART synthesizer.

\begin{table}[!htp]
\centering
\caption{RAPID estimates at default threshold $\tau = 0.3$ for the UCI Adult dataset ($n = 32{,}561$). Each row corresponds to one synthetic replicate generated via CART. Confidence intervals (95\%) obtained via non-parametric bootstrap ($R = 500$).}
\label{tab:rapid-adult}
\renewcommand{\arraystretch}{1.3}
\begin{tabular}{ccccc}
\toprule
 Replicate &  RAPID &  95\% CI (Lower) &  95\% CI (Upper) &  $n$ at risk \\
\midrule
 1 &  0.719 &  0.715 &  0.725 &  23,453 \\
 2 &  0.720 &  0.714 &  0.723 &  23,454 \\
 3 &  0.722 &  0.717 &  0.727 &  23,513 \\
 4 &  0.720 &  0.715 &  0.725 &  23,447 \\
 5 &  0.726 &  0.720 &  0.730 &  23,645 \\
\bottomrule
\end{tabular}
\end{table}

Is 72\% attribute-inference risk ``too high''? The answer is context-dependent. The Adult dataset exhibits strong predictive structure: income category is well-predicted by education, occupation, hours worked, and other quasi-identifiers. Any synthesizer that preserves this structure for analytical utility will inevitably also preserve the relationships that enable inference attacks. Datasets with weaker covariate--outcome relationships will typically yield lower RAPID values. Whether a given risk level is acceptable depends on the sensitivity of the attribute, the intended use case, and institutional risk tolerance. A data custodian observing high RAPID might consider several responses:
(i) applying a different synthesizer with stronger privacy guarantees,
(ii) suppressing or coarsening quasi-identifiers that drive predictive accuracy
(see Section~\ref{qi-attribution} for diagnostic tools),
(iii) restricting access to the synthetic data, or
(iv) accepting the risk for use cases where the sensitive attribute is not particularly confidential. RAPID does not prescribe a universal threshold for acceptability; rather, it provides the empirical foundation for informed, context-specific release decisions.

\section{Simulation study}
\label{sec:simulations}

\subsection{Overview and design}

To validate RAPID and investigate factors influencing attribute-inference
risk, we conducted three simulation studies:
\begin{enumerate}
\item \textbf{Dependency strength:} How does disclosure risk scale with the strength of the relationship between quasi-identifiers (QIs) and sensitive attributes? ($\kappa \in [0,100]$)

\item \textbf{Threshold sensitivity:} How does $\tau$ affect risk across
dependency regimes? (5 $\kappa$ levels $\times$ 19 $\tau$ values)

\item \textbf{QI attribution:} Which quasi-identifiers drive risk?
(Regression-based analysis)

\end{enumerate}

All simulations use synthetic health microdata with six variables (gender, age, education, income, health score, and disease status). We control dependency strength via a global parameter $\kappa \geq 0$, with full details in
~\ref{app:datagen}.

\subsection{Data generation process}

We simulate $n = 1{,}000$ independent records with six variables: gender ($G$), age ($A$), education ($E$), income ($I$), health score ($H$), and disease status ($D$), where $D$ serves as the sensitive attribute. The design encodes realistic dependencies through a latent socioeconomic
status (SES) variable, with dependency strength controlled by parameter $\kappa \geq 0$.

Signal and noise weights are derived from $\kappa$ as:
\[
w_{\text{signal}} = \sqrt{\frac{\kappa}{1+\kappa}}, \quad
w_{\text{noise}} = \sqrt{\frac{1}{1+\kappa}}.
\]

At $\kappa=0$, relationships are purely noise-driven ($w_{\text{signal}} = 0$); at $\kappa \gg 1$, dependencies approach deterministic strength ($w_{\text{signal}} \to 1$). The variable age follows a truncated normal; education is ordinal derived from a latent variable; income is log-linear; health score uses sigmoid transformation; disease status is generated via
multinomial logit with $\kappa$-scaled coefficients; gender is binary with mild SES dependency. Complete mathematical specifications appear in Appendix~\ref{app:datagen}.

\subsection{Dependency strength}

We varied dependency parameter $\kappa$ from 0 to 100 (101 values, 10 replications each) to investigate how attribute-inference risk scales with quasi-identifier--sensitive attribute relationship strength. Subplot (a) in Figure~\ref{fig:sim1_2} shows an S-shaped trajectory: risk escalates rapidly when transitioning from weak to moderate dependencies, then saturates as relationships approach deterministic strength. Saturation near 0.97 rather than 1.0 reflects inherent noise from the CART synthesizer's sampling. Attacker accuracy follows a similar trajectory (0.70 to 0.98), demonstrating that RAPID reliably reflects actual prediction performance: datasets with high RAPID face genuinely elevated disclosure risk.

Having established that risk scales with dependency strength, we next investigate how the threshold parameter $\tau$ interacts with this dependency structure. Simulation 2 addresses this challenge by examining threshold sensitivity across the full dependency spectrum.

\subsection{Threshold sensitivity}

\begin{figure}[!htp]
    \centering
    \includegraphics[width=1\linewidth]{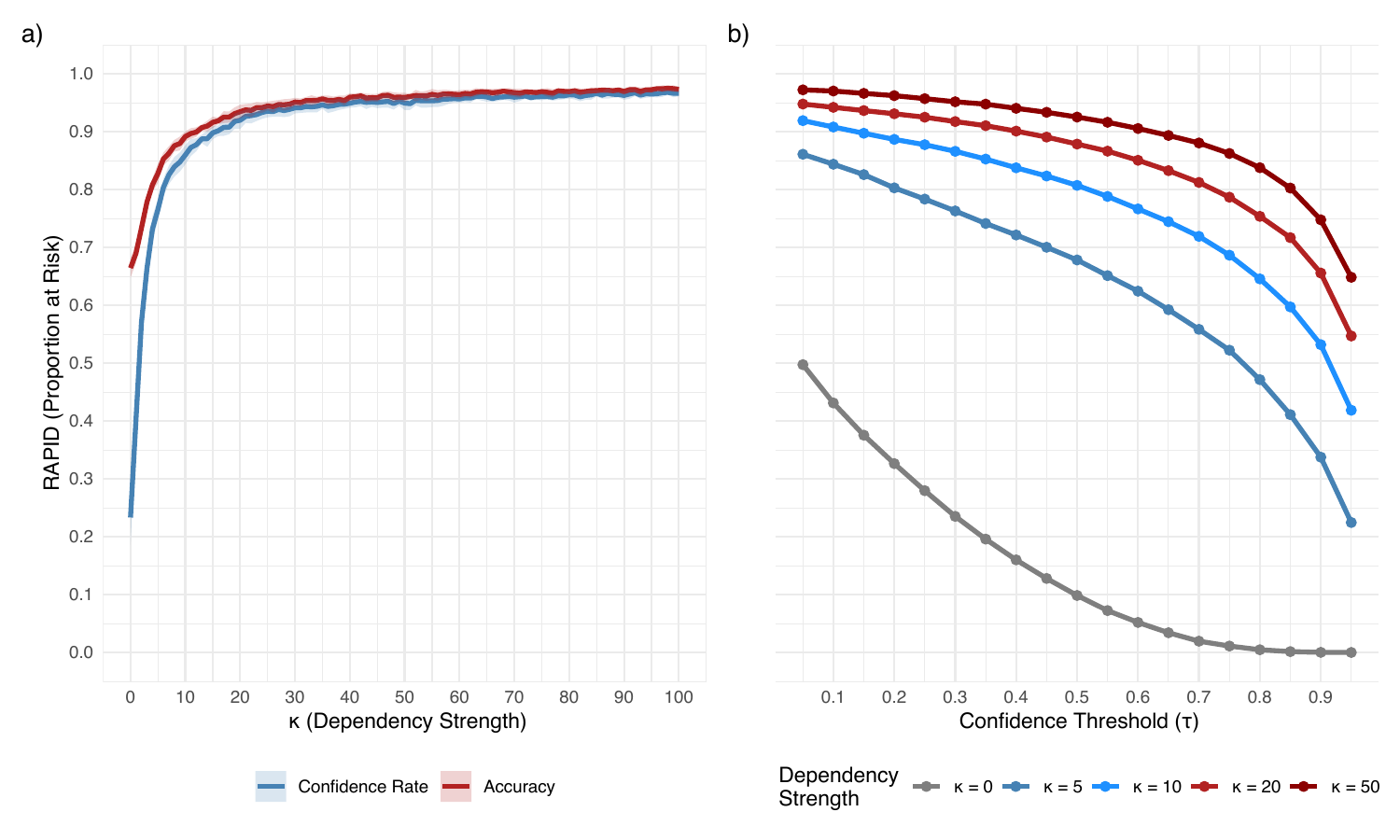}
    \caption{Impact of dependency strength and normalized gain threshold on RAPID:
(a) RAPID and attacker accuracy increase monotonically with dependency
strength $\kappa$. RAPID rises from 0.25 at $\kappa=0$ to 0.97 at $\kappa=100$,
with steepest increases at low $\kappa$ values ($\tau=0.3$). This
S-shaped growth demonstrates that attribute-inference risk escalates rapidly
when transitioning from weak to moderate quasi-identifier--sensitive attribute
relationships, then saturates as dependencies approach deterministic levels.
(b) RAPID vs.\ normalized gain threshold $\tau$ for varying $\kappa$. At low
dependency ($\kappa=0$, gray), the curve is convex, reflecting diffuse
attacker confidence where most records are filtered out at moderate thresholds.
At high dependency ($\kappa \geq 5$, blue/red), curves become concave, remaining
elevated until stringent thresholds ($\tau > 0.7$) are applied. This transition
reflects a qualitative shift in attacker confidence distributions as dependencies
strengthen.
Both panels: Mean $\pm$ 1 SD over 10 simulations; $n=1000$ records,
CART synthesizer, Random Forest attacker.}
    \label{fig:sim1_2}
\end{figure}

We examined how the normalized gain threshold $\tau$ affects RAPID across five dependency levels ($\kappa \in \{0, 5, 10, 20, 50\}$), varying $\tau$ from 0.05 to 0.95 in 0.05 increments (10 replications per combination). Subplot (b) in Figure~\ref{fig:sim1_2} reveals a qualitative shift in curve geometry: at low dependency ($\kappa=0$, gray), RAPID decreases convexly, dropping rapidly from 0.50 to near-zero by $\tau=0.60$, indicating diffuse attacker confidence. At high dependency ($\kappa \geq 5$, blue/red), curves remain concave, staying elevated until stringent thresholds ($\tau > 0.70$), reflecting concentrated confidence distributions near certainty. This convex-to-concave transition marks a shift from weak to strong adversarial inference capability.

Practically, the choice of $\tau$ should be guided by the sensitivity of the attribute rather than the proportion of flagged records: stricter thresholds ($\tau \geq 0.70$) are appropriate for highly sensitive attributes where only near-certain inferences constitute unacceptable risk, while lower thresholds ($\tau \approx 0.30$--$0.40$) provide broader protection when moderately confident inferences also pose disclosure concerns.
The threshold-dependency interaction reveals that the same $\tau$ corresponds to different levels of adversarial certainty depending on data structure, requiring calibration to the specific use case.

\subsection{Quasi-identifier attribution}
\label{qi-attribution}

\begin{figure}[!htp]
\hspace{-1cm}
    \includegraphics[width=1.1\linewidth]{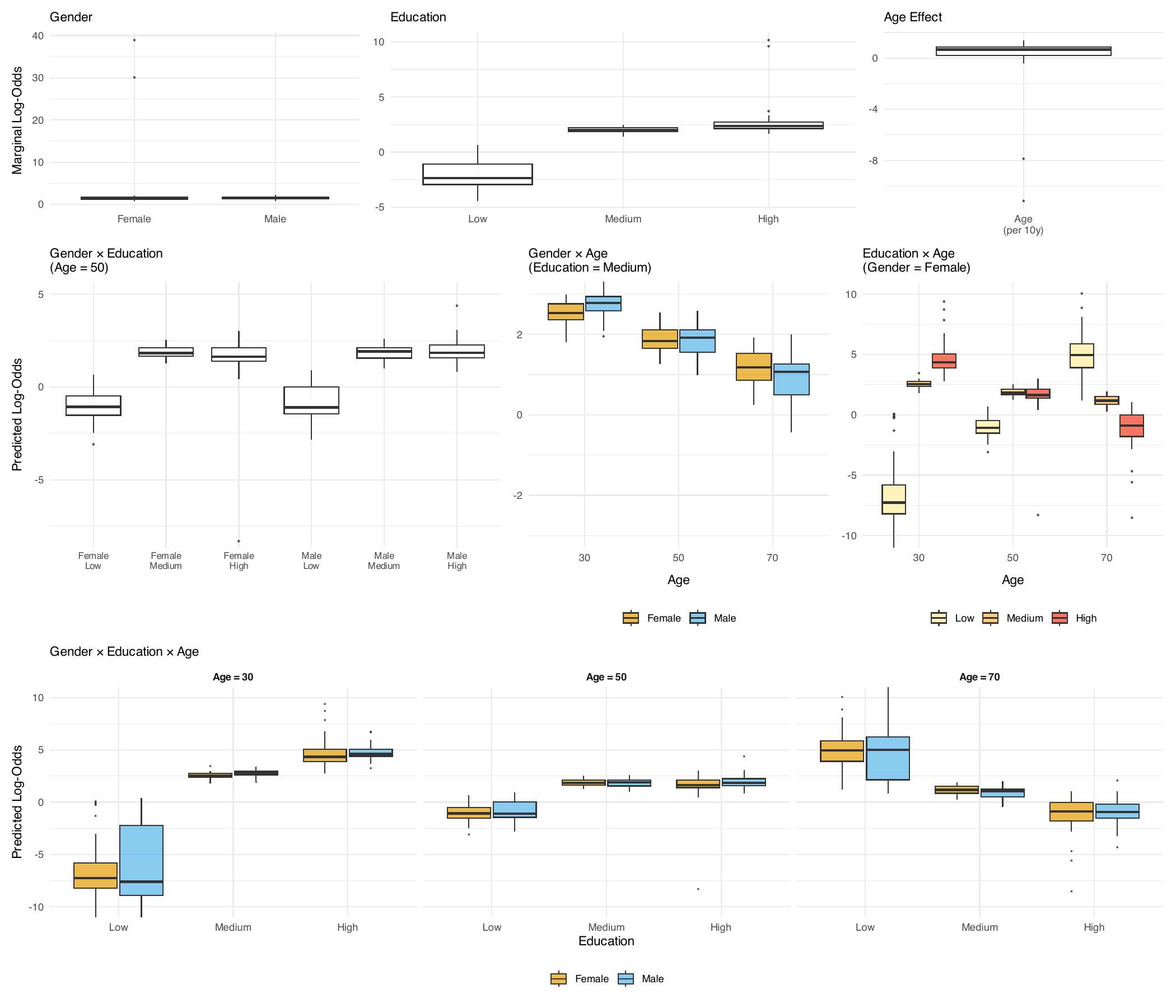}
 \caption{Quasi-identifier attribution analysis. 
Predicted log-odds of attribute inference risk from logistic regression 
across 50 simulations ($\kappa=10$, $\tau=0.3$, $n=1000$). 
\textbf{Top:} Marginal effects. 
\textbf{Middle:} Two-way interactions at specified conditioning values. 
\textbf{Bottom:} Three-way interaction at three age levels (30, 50, 70). 
Y-axes individually trimmed to enhance visibility; some extreme outliers not shown.}
    \label{fig:sim3}
\end{figure}

Figure~\ref{fig:sim3} illustrates the resulting predicted log-odds across demographic combinations. The analysis reveals strong associations between education and attribute inference risk, with substantial interaction effects--particularly between age and education levels--reflecting the strong quasi-identifier dependencies induced by the simulation design ($\kappa=10$).

Beyond these specific patterns, the example demonstrates RAPID's methodological advantage: row-wise risk flags enable attribution of inference vulnerability to specific quasi-identifier combinations through standard regression modeling. In real applications, such analysis could identify high-risk demographic groups for targeted disclosure control, test hypotheses about vulnerability drivers (e.g., does educational attainment moderate age-related risk?), and quantify effect sizes to prioritize mitigation efforts.

\newpage
\section{Discussion}
\label{sec:discussion}

RAPID provides a model-based, attacker-realistic approach to assessing attribute inference risk in released data. In this section, we discuss how RAPID relates to existing disclosure risk measures, address limitations, and offer practical guidance.

\subsection{Relationship to existing disclosure measures}

\begin{itemize}
\item \textbf{Match-based evaluation.}
RAPID shares the goal of quantifying attribute inference risk with measures like DCAP \citep{taub18} and DiSCO \citep{Raab_2025_Practical}, but differs fundamentally in approach. DiSCO evaluates risk by examining contingency tables: records are flagged disclosive when synthetic data within a quasi-identifier cell reveals unanimous sensitive attribute values, indicating that an attacker knowing the quasi-identifiers could infer the sensitive attribute with high confidence. DCAP similarly assess how accurately an attacker can predict sensitive attributes, but quantify risk via the attacker’s expected prediction accuracy, computed from the synthetic conditional target distributions rather than from an explicit classifier.

RAPID diverges from these approaches in three key aspects. First, it operates through predictive modeling rather than contingency table analysis, enabling natural handling of mixed-type data without requiring discretization of continuous variables or explicit enumeration of quasi-identifier cells. Second, RAPID's threshold-based risk criterion is tunable: for categorical sensitive attributes, data curators calibrate the normalized gain threshold $\tau$ to reflect attribute sensitivity and acceptable disclosure risk, while for continuous attributes, the threshold $\varepsilon$ specifies the acceptable percentage relative prediction error. Third, by incorporating baseline predictability directly into a record-level normalized gain criterion, RAPID ensures that flagged records represent genuine inference advantages beyond what marginal distributions alone would permit -- addressing the class imbalance problem where rare outcomes might appear ``perfectly predicted'' in small cells simply due to low base rates.

These design choices reflect RAPID's focus on adversarial realism: it models what a sophisticated attacker with machine learning capabilities could achieve, rather than relying on combinatorial uniqueness criteria. However, this approach introduces significant limitations. The reliance on machine learning models makes RAPID's risk assessments opaque and difficult to audit: unlike table-based methods where a data subject can verify whether their quasi-identifier cell contains unanimous attribute values, RAPID's predictions emerge from complex ensemble models. This opacity complicates stakeholder communication and regulatory compliance, particularly when data subjects exercise their right to explanation under privacy regulations. Moreover, RAPID's risk estimates are contingent on modeling choices -- attacker model type, hyperparameters, and the threshold parameters ($\tau$ for categorical, $\varepsilon$ for continuous attributes) -- introducing degrees of freedom that require justification but lack established guidelines. The thresholds $\tau$ and $\epsilon$, while flexible, are also arbitrary: there is no principled method for selecting $\tau$ or $\varepsilon$ beyond domain expertise and sensitivity judgments, the same threshold values correspond to vastly different levels of adversarial certainty depending on data structure. 
To explore how risk estimates vary with threshold choice, the \texttt{rapid::plot()} method applied on the \texttt{rapid::rapid()} result generates sensitivity curves showing the proportion of records flagged as at-risk across different values of $\tau$ (categorical) or $\varepsilon$ (continuous). See~\ref{app:r-code} for usage examples.

RAPID is suited for scenarios with complex quasi-identifier spaces, mixed-type data, or where modeling realistic attacker capabilities is paramount. Other table-based measures are preferable when transparency and alignment with classical SDC frameworks are priorities, or when regulatory or institutional requirements demand interpretable risk assessments. For critical applications, using both approaches in tandem may provide complementary perspectives on disclosure risk.

\item \textbf{Holdout-based evaluation.}
Holdout-based frameworks assess privacy via distance-to-closest-record (DCR), measuring whether synthetic records are closer to training than holdout records. While useful for detecting memorization, DCR does not quantify attribute inference risk (e.g., \citeauthor{Yao_2026_DCR}, \citeyear{Yao_2026_DCR}): high attribute inference risk can coexist with ``safe'' DCR values if the synthesizer preserves predictive relationships without copying records.

RAPID directly addresses attribute inference by training on released data and scoring on original covariates. The paradigm of training models on synthetic data and evaluating on original data is well established for \emph{utility} assessment -- measuring whether synthetic data preserve analytical relationships. RAPID repurposes this paradigm for \emph{risk} assessment: the same predictive accuracy that signals high utility also signals high disclosure risk when the prediction target is a sensitive attribute. Unlike DCR, RAPID provides an interpretable, bounded metric calibrated to class prevalence. 
Importantly, the R implementation of RAPID provided here supports both evaluation modes: the default match-based approach (training on synthetic data, scoring on all original records) and an optional holdout-based approach (training on synthetic data, scoring only on a holdout subset of original records not used to train the synthesizer). The holdout mode allows practitioners to assess both memorization risk and attribute inference risk within a unified framework. See Section~\ref{app:r-code} for implementation details.

\item \textbf{Bayesian approaches}
Bayesian disclosure risk measures \citep{Reiter_2014_Bayesiana, Hu_2021_Bayesian} provide posterior probabilities of attribute disclosure by integrating over uncertainty in the data-generating process. These approaches offer principled worst-case bounds but require distributional assumptions and can be computationally intensive.

RAPID takes a frequentist, simulation-based approach that is computationally efficient and assumption-light. Rather than providing worst-case bounds, RAPID estimates realized risk under a specified attacker model. For conservative assessments, practitioners can evaluate multiple attacker models and report $\mathrm{RAPID}_{\max}$ (Section~\ref{sec:rapid}).
\end{itemize}

\subsection{Limitations}
Several limitations should be acknowledged:

\begin{itemize}
\item \textbf{Threshold selection:} The choice of $\tau$ (categorical) and $\varepsilon$ (continuous) substantially affects risk estimates and requires domain expertise. While threshold choice is not unique to RAPID -- similar parameters appear in $k$-anonymity, $\ell$-diversity, DiSCO's unanimity criterion, and differential privacy's $\varepsilon$ -- there is no universally applicable guidance. The \texttt{rapid::plot()} method generates sensitivity curves to explore how flagged record proportions vary with threshold choice (see Appendix~\ref{app:r-code}). For applications where thresholds are difficult to justify, data-driven approaches such as permutation tests offer an alternative.

\item \textbf{Model dependence:} RAPID's risk estimates depend on the choice of attacker model. Different models may produce different risk assessments for the same record. This is both a feature -- capturing that different attackers pose different threats -- and a limitation, as it introduces ambiguity. For conservative assessments, practitioners can evaluate multiple attacker models and report $\mathrm{RAPID}_{\max}$, the maximum risk across models. The inherent opacity of machine learning models makes RAPID's assessments harder to audit than relying on combinatorial uniqueness criteria that may underestimate risk. This further complicates regulatory compliance and stakeholder communication.

\item \textbf{Single sensitive attribute:} The current formulation assesses risk for one sensitive attribute at a time. In practice, datasets often contain multiple sensitive variables. Extending RAPID to joint or sequential assessment of multiple attributes is a direction for future work.
\end{itemize}

\subsection{Practical guidance}

Based on our empirical results, we offer the following guidance for practitioners:
\begin{itemize}
\item Use $\tau = 0.3$ as a starting point for categorical attributes and $\varepsilon = 10\%$ for continuous attributes. This choice is very variable dependent, for high sensitive variables, consider stricter thresholds.
\item Evaluate RAPID using multiple attacker models (e.g., random forest, gradient boosting) and report the maximum observed risk for a conservative assessment.
\item Generate threshold curves (RAPID vs.\ $\tau$) using the \texttt{rapid::plot()} method to understand sensitivity and guide threshold selection for specific policy requirements.
\item Use per-record risk flags for attribution analysis: stratify RAPID by subgroups, cross-tabulate risk rates across QI combinations, or fit regression models to identify which quasi-identifier patterns drive disclosure risk.
\item Contextualize RAPID alongside utility metrics to navigate the privacy-utility trade-off when comparing synthesizers or anonymization strategies.

\item Apply mitigation strategies for high-risk records: When RAPID identifies unacceptable risk levels, several interventions are available. Do not simply remove high-risk records -- this shifts the risk distribution and creates new high-risk records in subsequent evaluations. Instead, consider: 
\begin{enumerate}
\item \emph{Reduce quasi-identifier granularity} through coarsening (binning continuous variables, merging categories) or formal anonymization methods. Use attribution analysis (Section~\ref{qi-attribution}) to identify which quasi-identifiers drive predictive accuracy;
\item \emph{Exclude highly predictive quasi-identifiers} from the released data if not essential for analytical utility;
\item \emph{Add post-processing noise} to sensitive attributes in the synthetic data;
\item \emph{Retrain the synthesizer} with stronger privacy constraints (e.g., differential privacy, restricted conditional sampling);
\item \emph{Restrict access} to the synthetic data through controlled environments or data use agreements;
\item \emph{Accept the risk} for use cases where the sensitive attribute is not highly confidential and stakeholders understand the trade-offs.
\end{enumerate}
\end{itemize}

\section{Conclusion}
\label{sec:conclusion}

Classical statistical disclosure control focused on identity disclosure and record uniqueness, but modern synthetic data generators pose a different threat: high-fidelity synthesis preserves predictive structure so well that sensitive attributes may be inferable from quasi-identifiers alone. RAPID addresses this challenge by directly measuring attribute inference vulnerability under realistic attacker assumptions. Where previous measures asked ``
whether SDGs replicated identifying and sensitive information?'', RAPID asks ``can an attacker confidently predict what they should not know?''

This reframing enables practical decisions that were previously difficult. Data custodians can now compare synthesizers not only on utility but on inferential risk, identifying which methods leak predictive signal and which preserve privacy. The per-record risk scores reveal high-risk subpopulations -- combinations of quasi-identifiers where inference succeeds -- enabling targeted remediation rather than blanket suppression. Perhaps most importantly, RAPID helps answer the question implicit in every high-utility synthetic release: is the utility \emph{too} high, reflecting dangerous fidelity to the original data's predictive structure?

We emphasize what RAPID deliberately does not claim. It is not a privacy guarantee in the formal sense of differential privacy, nor does it provide worst-case bounds or eliminate the possibility of inferential disclosure. Rather, RAPID is a scenario-based empirical diagnostic: it quantifies realized vulnerability under a specified attacker model, conditional on the predictive structure preserved in the release. This positions RAPID as a complement to -- not a replacement for -- formal privacy frameworks. Where differential privacy offers provable guarantees at the cost of utility and interpretability, RAPID offers transparent, actionable risk assessment that supports informed release decisions.

The timing of this contribution reflects broader trends in data sharing. Open science mandates increasingly require data availability, yet legal and ethical constraints on microdata release remain. Synthetic data have emerged as a pragmatic solution, with adoption accelerating across official statistics, health research, and social science. Simultaneously, advances in machine learning and the growing availability of public information about individuals have substantially increased the feasibility of privacy attacks.
RAPID provides the missing diagnostic: a way to assess whether synthetic data that \emph{look} safe actually \emph{are} safe against inference attacks that modern ML makes trivial to mount.

Future work includes extensions to multiple sensitive attributes, integration with utility metrics in unified risk-utility dashboards, and development of iterative workflows where high-risk records identified by RAPID are selectively protected before re-evaluation. The RAPID framework applies in principle to longitudinal data, where temporal features provide additional quasi-identifiers and risk can be assessed per observation or aggregated per individual; adapting the methodology to account for within-individual correlation is a natural direction for further research.

\section*{Code Availability}

The core RAPID \cite{RAPID} implementation is available at 
\url{https://github.com/qwertzlbry/RAPID}. 
Complete reproducibility code including all simulation scripts 
(Section~\ref{sec:simulations}) and the real data analysis 
(Section~\ref{sec:adult}) will be released upon journal publication.

\section*{Acknowledgments}
This work was supported by the Swiss National Science Foundation (SNSF) Bridge Discovery grant no.\ 211751.
We thank Jiří Novák for his contributions to the early implementation of RAPID and discussions.

\bibliography{references}

@InProceedings{taub18,
author="Taub, Jennifer and Elliot, Mark and Pampaka, Maria and Smith, Duncan",
editor="Domingo-Ferrer, Josep
and Montes, Francisco",
title="Differential Correct Attribution Probability for Synthetic Data: An Exploration",
booktitle="Privacy in Statistical Databases",
year="2018",
publisher="Springer International Publishing",
address="Cham",
pages="122–137",
isbn="978-3-319-99771-1",
url={https://doi.org/10.1007/978-3-319-99771-1_9}
}

@manual{raab2024synthpop,
  author       = {Raab, Gillian M. and Nowok, Beata and Dibben, Chris and Snoke, Joshua},
  title        = {synthpop User Guide},
  year         = {2024},
  organization = {University of Edinburgh},
  note         = {R package vignette/manual; attribute disclosure tools incl.\ DiSCO},
  url          = {https://cran.r-project.org/package=synthpop}
}

@InProceedings{kwatra24,
author="Kwatra, Saloni
and Torra, Vicen{\c{c}}",
editor="Bieker, Felix
and de Conca, Silvia
and Gruschka, Nils
and Jensen, Meiko
and Schiering, Ina",
title="Empirical Evaluation of Synthetic Data Created by Generative Models via Attribute Inference Attack",
booktitle="Privacy and Identity Management. Sharing in a Digital World",
year="2024",
publisher="Springer Nature Switzerland",
address="Cham",
pages="282–291",
isbn="978-3-031-57978-3",
url={https://doi.org/10.1007/978-3-031-57978-3_18}
}

@misc{SNF_ORD2024,
  author       = {{Swiss National Science Fund (SNSF)}},
  title        = {{Open Research Data: Ein erster Blick auf die aktuelle Praxis}},
  year         = {2024},
  howpublished = {\url{https://data.snf.ch/stories/open-research-data-2023-de.html}},
  note         = {accessed October 15, 2025},
}

@inproceedings{ward25,
  title = {Ensembling Membership Inference Attacks Against Tabular Generative Models},
  booktitle = {Proceedings of the 18th {{ACM Workshop}} on {{Artificial Intelligence}} and {{Security}}},
  author = {Ward, Joshua and Yang, Yuxuan and Wang, Chi-Hua and Cheng, Guang},
  year = 2025,
  series = {{{AISec}} '25},
  pages = {182-193},
  publisher = {ACM},
  address = {New York, NY, USA},
  url = {https://doi.org/10.1145/3733799.3762977},
  isbn = {979-8-4007-1895-3}
}

@article{miletic24,
    author = {Miletic, Marko and Sariyar, Murat},
    title = {Challenges of Using Synthetic Data Generation Methods for Tabular Microdata},
    journal = {Applied Sciences},
    volume = {14},
    year = {2024},
    number = {14},
    pages = {Article 5975},
    issn = {2076-3417},
    url = {https://doi.org/10.3390/app14145975}
}

@misc{mekonnen24,
      title={Conditioning {GAN} Without Training Dataset}, 
      author={Kidist Amde Mekonnen},
      year={2024},
      eprint={2405.20687},
      archivePrefix={arXiv},
      primaryClass={cs.CV},
      url={https://doi.org/10.48550/arXiv.2405.20687}, 
}

@misc{Thees25pca,
    author = {Thees, Oscar and M\"uller, Roman and Templ, Matthias},
    title = {Beyond the Trade-off Curve: Multivariate and Advanced Risk-Utility Maps for Evaluating  Anonymized and Synthetic Data},
    journal = {Arxiv},
    year = {2025},
    url = {https://doi.org/10.48550/arXiv.2510.23500}
}

@article{simPop,
    author = {Templ, Matthias and Meindl, Bernhard and Kowarik, Alexander and Dupriez, Olivier},
    title = {Simulation of Synthetic Complex Data: The {R} Package sim{P}op},
    journal = {Journal of Statistical Software},
    year = {2017},
    volume = {79},
    number = {10},
    pages = {1–38},
    url = {https://doi.org/10.18637/jss.v079.i10},
}

@article{barrientos18,
author = {Andr{\'e}s F. Barrientos and Alexander Bolton and Tom Balmat and Jerome P. Reiter and John M. de Figueiredo and Ashwin Machanavajjhala and Yan Chen and Charley Kneifel and Mark DeLong},
title = {{Providing access to confidential research data through synthesis and verification: An application to data on employees of the U.S. federal government}},
volume = {12},
journal = {The Annals of Applied Statistics},
number = {2},
publisher = {Institute of Mathematical Statistics},
pages = {1124–1156},
year = {2018},
url = {https://doi.org/10.1214/18-AOAS1194}
}

@misc{MuralidharRuggles2024,
  author       = {Muralidhar, Krishnamurty and Ruggles, Steven},
  title        = {Escalation of Commitment: A Case Study of the United States Census Bureau Efforts to Implement Differential Privacy for the 2020 Decennial Census},
  year         = {2024},
  eprint       = {2407.15957},
  archivePrefix= {arXiv},
  primaryClass = {cs.CR},
  url          = {https://doi.org/10.48550/arXiv.2407.15957},
}

@article{BlancoJusticia2022,
  author    = {Blanco-Justicia, Alberto and S{\'a}nchez, David and Domingo-Ferrer, Josep and Muralidhar, Krishnamurty},
  title     = {A Critical Review on the Use (and Misuse) of Differential Privacy in Machine Learning},
  journal   = {ACM Computing Surveys},
  volume    = {55},
  number    = {8},
  pages     = {Article 160},
  year      = {2022},
  url       = {https://doi.org/10.1145/3547139},
  publisher = {ACM}
}

@article{domingo21,
    author = {Domingo-Ferrer, Josep and S\'{a}nchez, David and Blanco-Justicia, Alberto},
    title = {The limits of differential privacy (and its misuse in data release and machine learning)},
    year = {2021},
    publisher = {Association for Computing Machinery},
    address = {New York, NY, USA},
    volume = {64},
    number = {7},
    issn = {0001-0782},
    url = {https://doi.org/10.1145/3433638},
    journal = {Communications of the ACM},
    pages = {33–35},
    numpages = {3}
}

@article{muralidhar23,
    author = {Krishnamurty Muralidhar and Josep Domingo-Ferrer},
    title ={Database Reconstruction Is Not So Easy and Is Different from Reidentification},
    journal = {Journal of Official Statistics},
    volume = {39},
    number = {3},
    pages = {381–398},
    year = {2023},
    url = {https://doi.org/10.2478/jos-2023-0017}
}

@misc{francis2025betterattributeinferencevulnerability,
      title={Towards Better Attribute Inference Vulnerability Measures}, 
      author={Paul Francis and David Wagner},
      year={2025},
      eprint={2507.01710},
      archivePrefix={arXiv},
      primaryClass={cs.CR},
      url={https://doi.org/10.48550/arXiv.2507.01710},
}

@techreport{Duncan01a,
	Author = {Duncan, George and Keller-McNulty, Sallie and Stokes,  Lynne},
	Institution = {Los Alamos National Laboratory},
	Title = {Disclosure risk vs. data utility: the {R-U} confidentiality map},
	Type = {Technical Report {LA-UR-01-6428}},
	Url = {https://www.niss.org/sites/default/files/technicalreports/tr121.pdf},
	Year = {2001}
}

@article{PlatzerReutterer2021Holdout,
  author  = {Platzer, Michael and Reutterer, Thomas},
  title   = {Holdout-Based Empirical Assessment of Mixed-Type Synthetic Data},
  journal = {Frontiers in Big Data},
  year    = {2021},
  volume  = {4},
  pages   = {Article 679939},
  url     = {https://doi.org/10.3389/fdata.2021.679939},
}

@inproceedings{hittmeir20,
    author = {Hittmeir, Markus and Mayer, Rudolf and Ekelhart, Andreas},
    title = {A Baseline for Attribute Disclosure Risk in Synthetic Data},
    year = {2020},
    isbn = {9781450371070},
    publisher = {ACM},
    address = {New York, NY, USA},
    url = {https://doi.org/10.1145/3374664.3375722},
    booktitle = {Proceedings of the Tenth ACM Conference on Data and Application Security and Privacy},
    pages = {133–143},
    numpages = {11},
    location = {New Orleans, LA, USA},
    series = {CODASPY '20}
}

@article{templ14risk,
    title = {Providing Data with High Utility and no Disclosure Risk for the Public and Researchers: An Evaluation by Advanced Statistical Disclosure Risk},
    volume={43}, 
    author = {Templ, Matthias},
    url={https://doi.org/10.17713/ajs.v43i4.43},  
    number={4}, 
    year = {2014},
    journal={Austrian Journal of Statistics}, 
    pages={247–254} 
}

@article{emam20,
  author    = {El Emam, Khaled and Mosquera, Laura and Bass, Jason},
  title     = {Evaluating Identity Disclosure Risk in Fully Synthetic Health Data: Model Development and Validation},
  journal   = {Journal of Medical Internet Research},
  year      = {2020},
  volume    = {22},
  number    = {11},
  pages     = {Article e23139},
  url       = {https://doi.org/10.2196/23139},
  pmid      = {33196453},
  pmcid     = {PMC7704280}
}

@book{Hundepool_2012_Statistical,
    location = {Chichester, {UK}},
    edition = {1},
    title = {Statistical disclosure control},
    rights = {http://doi.wiley.com/10.1002/tdm\_license\_1.1},
    isbn = {978-1-119-97815-2},
    url = {https://doi.org/10.1002/9781118348239},
    series = {Wiley series in survey methodology},
    shorttitle = {Statistical disclosure control},
    pagetotal = {288},
    publisher = {John Wiley \& Sons, Ltd},
    author = {Hundepool, Anco and Domingo‐Ferrer, Josep and Franconi, Luisa and Giessing, Sarah and Nordholt, Eric Schulte and Spicer, Keith and de Wolf, Peter‐Paul},
    editorb = {Couper, Mick P. and Kalton, Graham and Lyberg, Lars and Rao, J.N.K. and Schwarz, Norbert and Skinner, Christopher},
    editorbtype = {redactor},
    urldate = {2025-03-04},
    year = {2012},
    langid = {english},
}

@article{Rubin93,
	Author = {Rubin, Donald B.},
	Journal = {Journal of Official Statistics},
	Number = {2},
	Pages = {461–468},
	Title = {Discussion of statistical disclosure limitation},
	Volume = {9},
	Year = {1993}}

@Manual{RAPID,
    title = {{RAPID}: Risk Assessment through Prediction for Inference-based Disclosure},
    author = {Oscar Thees and Matthias Templ and Roman Müller},
    note = {R package version 0.1.0},
    url = {https://github.com/qwertzlbry/RAPID},
    year = 2026
}

@InProceedings{thees24,
    author="Thees, Oscar
    and Nov{\'a}k, Ji{\v{r}}{\'i}
    and Templ, Matthias",
    editor="Domingo-Ferrer, Josep
    and {\"O}nen, Melek",
    title="Evaluation of Synthetic Data Generators on Complex Tabular Data",
    booktitle="Privacy in Statistical Databases",
    year="2024",
    publisher="Springer Nature Switzerland",
    address="Cham",
    pages="194–209",
    url={https://doi.org/10.1007/978-3-031-69651-0_13}
}

@misc{Raab_2025_Practical,
  title = {Practical Privacy Metrics for Synthetic Data},
  author = {Raab, Gillian M. and Nowok, Beata and Dibben, Chris},
  year = {2025},
  eprint = {2406.16826},
  eprinttype = {arXiv},
  pages = {1–23},
  url = {https://doi.org/10.48550/arXiv.2406.16826},
  urldate = {2025-08-25},
  langid = {english},
  pubstate = {prepublished},
}

@article{DomingoFerrer_2025_Statistical,
  title = {Statistical {{Disclosure Control}}: {{Moving Forward}}},
  shorttitle = {Statistical {{Disclosure Control}}},
  author = {{Domingo-Ferrer}, Josep and S{\'a}nchez, David and Muralidhar, Krishnamurty},
  year = 2025,
  journal = {Journal of Official Statistics},
  volume = {41},
  number = {3},
  pages = {820–826},
  issn = {0282-423X},
  url = {https://doi.org/10.1177/0282423X241312023},
  urldate = {2025-11-20},
  copyright = {https://creativecommons.org/licenses/by-nc/4.0/},
  langid = {english},
}

@article{Duncan_1989_Risk,
  title = {The Risk of Disclosure for Microdata},
  author = {Duncan, George and Lambert, Diane},
  year = 1989,
  journal = {Journal of Business \& Economic Statistics},
  volume = {7},
  number = {2},
  pages = {207–217},
  url = {https://doi.org/10.2307/1391438},
  langid = {english},
}

@article{Dalenius_1977_Methodology,
  title = {Towards a Methodology for Statistical Disclosure Control},
  author = {Dalenius, Tore},
  year = 1977,
  journal = {Statistisk Tidskrift},
  volume = {15},
  pages = {429–444},
  issn = {0039-7261}
}

@article{2015_Nosek,
	title = {Promoting an open research culture},
	volume = {348},
	issn = {0036-8075},
	url = {https://doi.org/10.1126/science.aab2374},
	pages = {1422–1425},
	number = {6242},
	journal = {Science},
	author = {Nosek, B. A. and Alter, G. and Banks, G. C. and Borsboom, D. and Bowman, S. D. and Breckler, S. J. and Buck, S. and Chambers, C. D. and Chin, G. and Christensen, G. and Contestabile, M. and Dafoe, A. and Eich, E. and Freese, J. and Glennerster, R. and Goroff, D. and Green, D. P. and Hesse, B. and Humphreys, M. and Ishiyama, J. and Karlan, D. and Kraut, A. and Lupia, A. and Mabry, P. and Madon, T. and Malhotra, N. and Mayo-Wilson, E. and {McNutt}, M. and Miguel, E. and Levy Paluck, E. and Simonsohn, U. and Soderberg, C. and Spellman, B. A. and Turitto, J. and {VandenBos}, G. and Vazire, S. and Wagenmakers, E. J. and Wilson, R. and Yarkoni, T.},
	year = {2015},
	pmid = {26113702},
	pmcid = {PMC4550299},
}

@article{Muhlhoff_2021_Predictive,
  title = {Predictive Privacy: Towards an Applied Ethics of Data Analytics},
  shorttitle = {Predictive Privacy},
  author = {M{\"u}hlhoff, Rainer},
  year = 2021,
  journal = {Ethics and Information Technology},
  volume = {23},
  number = {4},
  pages = {675–690},
  issn = {1388-1957, 1572-8439},
  url = {https://doi.org/10.1007/s10676-021-09606-x},
  urldate = {2026-01-13},
  langid = {english},
}

@article{Dwork_2010_Difficulties,
  title = {On the Difficulties of Disclosure Prevention in Statistical Databases or the Case for Differential Privacy},
  author = {Dwork, Cynthia and Naor, Moni},
  year = 2010,
  journal = {Journal of Privacy and Confidentiality},
  volume = {2},
  number = {1},
  pages = {93–107},
  issn = {2575-8527},
  url = {https://doi.org/10.29012/jpc.v2i1.585},
  urldate = {2025-12-16},
  langid = {english},
}

@article{Reiter_2014_Bayesiana,
  title = {Bayesian Estimation of Disclosure Risks for Multiply Imputed, Synthetic Data},
  author = {Reiter, Jerome P. and Wang, Quanli and Zhang, Biyuan},
  year = 2014,
  journal = {Journal of Privacy and Confidentiality},
  volume = {6},
  number = {1},
  issn = {2575-8527},
  url = {https://doi.org/10.29012/jpc.v6i1.635},
  urldate = {2026-01-14},
  copyright = {http://creativecommons.org/licenses/by-sa/4.0},
  langid = {english},
}

@misc{adult_2,
  author       = {Becker, Barry and Kohavi, Ronny},
  title        = {Adult [{Dataset}]},
  year         = {1996},
  howpublished = {UCI Machine Learning Repository},
  url          = {https://doi.org/10.24432/C5XW20}
}

@book{Willenborg_2001_Elements,
	location = {New York, {NY}, {USA}},
	edition = {1},
	title = {Elements of statistical disclosure control},
	volume = {155},
	rights = {http://www.springer.com/tdm},
	isbn = {978-1-4613-0121-9},
	series = {Lecture Notes in Statistics},
	pagetotal = {261},
	publisher = {Springer},
	author = {Willenborg, Leon and de Waal, Ton},
	editorb = {Bickel, P. and Diggle, P. and Fienberg, S. and Krickeberg, K. and Olkin, I. and Wermuth, N. and Zeger, S.},
	editorbtype = {redactor},
	urldate = {2025-03-14},
	year = {2001},
	langid = {english},
	url = {https://doi.org/10.1007/978-1-4613-0121-9}
}

@article{Nowok_2016_Synthpop,
	title = {synthpop: Bespoke creation of synthetic data in {R}},
	volume = {74},
	issn = {1548-7660},
	url = {https://doi.org/10.18637/jss.v074.i11},
	shorttitle = {synthpop},
	pages = {1–26},
	number = {11},
	journal = {Journal of Statistical Software},
	shortjournal = {{JSS}},
	author = {Nowok, Beata and Raab, Gillian M. and Dibben, Chris},
	urldate = {2025-05-23},
	year = {2016},
	langid = {english}
}

@inproceedings{Dwork_2006_Calibrating,
  title = {Calibrating Noise to Sensitivity in Private Data Analysis},
  booktitle = {Proceedings of the {{Third}} Conference on {{Theory}} of {{Cryptography}}},
  author = {Dwork, Cynthia and McSherry, Frank and Nissim, Kobbi and Smith, Adam},
  year = 2006,
  pages = {265–284},
  publisher = {Springer-Verlag},
  address = {Berlin, Heidelberg},
  url = {https://doi.org/10.1007/11681878_14},
  langid = {english}
}

@article{Alfons_2011_Simulation,
  title = {Simulation of Close-to-Reality Population Data for Household Surveys with Application to {{EU-SILC}}},
  author = {Alfons, Andreas and Kraft, Stefan and Templ, Matthias and Filzmoser, Peter},
  year = 2011,
  journal = {Statistical Methods \& Applications},
  volume = {20},
  number = {3},
  pages = {383–407},
  issn = {1618-2510, 1613-981X},
  url = {https://doi.org/10.1007/s10260-011-0163-2},
  urldate = {2026-01-28},
  langid = {english}
}

@article{Wright_2017_Ranger,
  title = {{ranger}: A Fast Implementation of Random Forests for High Dimensional Data in {{C}}++ and {{R}}},
  author = {Wright, Marvin N. and Ziegler, Andreas},
  year = 2017,
  journal = {Journal of Statistical Software},
  volume = {77},
  number = {1},
  pages = {1–17},
  url = {https://doi.org/10.18637/jss.v077.i01}
}

@inproceedings{Chen_2016_XGBoost,
  title = {{{XGBoost}}: {{A Scalable Tree Boosting System}}},
  booktitle = {Proceedings of the 22nd {{ACM SIGKDD International Conference}} on {{Knowledge Discovery}} and {{Data Mining}}},
  author = {Chen, Tianqi and Guestrin, Carlos},
  year = 2016,
  month = aug,
  pages = {785–794},
  publisher = {ACM},
  address = {San Francisco California USA},
  url = {https://doi.org/10.1145/2939672.2939785},
  urldate = {2026-01-29},
  isbn = {978-1-4503-4232-2},
  langid = {english}
}

@article{wilson1927probable,
  author = {Wilson, Edwin B.},
  title = {Probable Inference, the Law of Succession, and Statistical Inference},
  journal = {Journal of the American Statistical Association},
  volume = {22},
  number = {158},
  pages = {209--212},
  year = {1927},
  url = {https://doi.org/10.1080/01621459.1927.10502953}
}

@article{clopper1934use,
  author = {Clopper, Charles J. and Pearson, Egon S.},
  title = {The Use of Confidence or Fiducial Limits Illustrated in the Case of the Binomial},
  journal = {Biometrika},
  volume = {26},
  number = {4},
  pages = {404--413},
  year = {1934},
  url = {https://doi.org/10.2307/2331986}
}

@article{Hu_2021_Bayesian,
  author = {Hu, Jingchen and Savitsky, Terrance D. and Williams, Matthew R.},
  title = {Bayesian Estimation of Population Size and Overlap from Random Samples},
  journal = {Journal of Privacy and Confidentiality},
  volume = {11},
  number = {1},
  year = {2021},
  url = {https://doi.org/10.29012/jpc.748}
}

@inproceedings{Slokom_2022_When,
  title = {When Machine Learning Models Leak: An Exploration of Synthetic Training Data},
  shorttitle = {When Machine Learning Models Leak},
  booktitle = {Privacy in {{Statistical Databases}}},
  author = {Slokom, Manel and {de Wolf}, Peter-Paul and Larson, Martha},
  editor = {{Domingo-Ferrer}, Josep and Laurent, Maryline},
  year = 2022,
  series = {Lecture {{Notes}} in {{Computer Science}}},
  volume = {13463},
  pages = {283--296},
  publisher = {Springer},
  address = {Cham, Switzerland},
  issn = {0302-9743},
  url = {https://doi.org/10.1007/978-3-031-13945-1_20},
  isbn = {978-3-031-13945-1},
  langid = {english}
}

@InProceedings{Yao_2026_DCR,
author="Yao, Zexi
and Kr{\v{c}}o, Nata{\v{s}}a
and Ganev, Georgi
and de Montjoye, Yves-Alexandre",
editor="Nicomette, Vincent
and Benzekri, Abdelmalek
and Boulahia-Cuppens, Nora
and Vaidya, Jaideep",
title="The DCR Delusion: Measuring the Privacy Risk of Synthetic Data",
booktitle="Computer Security -- ESORICS 2025",
year="2026",
publisher="Springer Nature Switzerland",
address="Cham",
pages="469--487",
url = {https://doi.org/10.1007/978-3-032-07884-1_24}
}

\appendix

\section{Simulation data generation}
\label{app:datagen}

We simulate $n$ independent microdata records with six variables: gender ($G$), age ($A$), education ($E$), income ($I$), health score ($H$), and disease status ($D$). The design encodes realistic, policy-relevant dependencies through a latent socioeconomic status (SES) variable while remaining transparent and tunable via a global dependency parameter $\kappa\ge 0$. Throughout, $\mathsf{TN}(\mu,\sigma^2;[a,b])$ denotes a normal distribution truncated to $[a,b]$, and we use standardized predictors $\tilde{X}=(X-\mu_X)/\sigma_X$ when indicated.

\paragraph{Dependency mechanism.}
All variable dependencies flow through a shared latent variable:
\[
\text{SES}_i \sim \mathcal{N}(0,1),
\]
representing unobserved socioeconomic status. The strength of dependencies is controlled by signal and noise weights derived from $\kappa$:
\[
w_{\text{signal}} = \sqrt{\frac{\kappa}{1+\kappa}}, \qquad
w_{\text{noise}} = \sqrt{\frac{1}{1+\kappa}}.
\]
At $\kappa=0$, relationships are driven purely by noise ($w_{\text{signal}}=0$, $w_{\text{noise}}=1$). As $\kappa\to\infty$, dependencies become deterministic ($w_{\text{signal}}\to 1$, $w_{\text{noise}}\to 0$). At the default $\kappa=1$, signal and noise contribute equally ($w_{\text{signal}}=w_{\text{noise}}=1/\sqrt{2}$), yielding approximately 50\% explained variance.

\paragraph{Age.}
We draw age from a truncated normal to reflect adult populations:
\[
A_i \sim \mathsf{TN}(\mu_A,\sigma_A^2;[a_{\min},a_{\max}]),
\quad \text{defaults: } \mu_A=45,\ \sigma_A=12,\ a_{\min}=18,\ a_{\max}=85.
\]

\paragraph{Education.}
Education is an ordinal categorical variable with three levels $\{0,1,2\}$ corresponding to \emph{low}, \emph{medium}, and \emph{high} attainment. We generate $E_i$ from a latent continuous variable that depends on both SES and age:
\[
L_i = w_{\text{signal}} \cdot (0.8\cdot\text{SES}_i - 0.4\cdot\tilde{A}_i) + w_{\text{noise}}\cdot\varepsilon_i,
\quad \varepsilon_i\sim\mathcal{N}(0,1),
\]
\[
E_i =
\begin{cases}
0 & \text{if } L_i < c_1, \\
1 & \text{if } c_1 \le L_i < c_2, \\
2 & \text{if } L_i \ge c_2,
\end{cases}
\quad c_1 = -0.3,\ c_2 = 0.7.
\]
The negative age coefficient reflects the empirical pattern of younger cohorts having higher educational attainment.

\paragraph{Income.}
We generate log-income using a linear model:
\[
\log I_i^\star = w_{\text{signal}}\cdot(0.5\cdot\text{SES}_i + 0.3\cdot\tilde{A}_i + 0.25\cdot E_i) + w_{\text{noise}}\cdot\varepsilon_i,
\quad \varepsilon_i\sim\mathcal{N}(0,1),
\]
\[
I_i = \exp(10 + \log I_i^\star).
\]
The constant 10 centers income around realistic values (approximately \$20,000--\$40,000).

\paragraph{Health score.}
$H$ is a continuous health measure on $[0,100]$ created via sigmoid transformation of a linear predictor:
\[
H_i^\star = w_{\text{signal}}\cdot\bigl(0.6\cdot\text{SES}_i - 0.5\cdot\tilde{A}_i + 0.2\cdot E_i + 0.2\cdot\widetilde{\log I_i}\bigr) + w_{\text{noise}}\cdot\zeta_i,
\quad \zeta_i\sim\mathcal{N}(0,1),
\]
\[
H_i = \frac{100}{1 + \exp(-H_i^\star)}.
\]
The negative age coefficient reflects declining health with age, while positive SES, education, and income coefficients represent protective effects.

\paragraph{Disease status.}
$D_i\in\{\texttt{healthy},\texttt{diabetic},\texttt{hypertensive}\}$ is sampled via a multinomial logit with \texttt{healthy} as the baseline. Unlike other variables, disease dependencies scale \emph{linearly} with $\kappa$ rather than through signal/noise weights:
\begin{align*}
\log\frac{\Pr(D_i=\texttt{diabetic})}{\Pr(D_i=\texttt{healthy})}
&= -1.5 + \kappa\cdot\bigl(0.8\cdot\tilde{A}_i - 0.3\cdot\widetilde{\log I_i} - 0.2\cdot E_i\bigr),\\
\log\frac{\Pr(D_i=\texttt{hypertensive})}{\Pr(D_i=\texttt{healthy})}
&= -1.3 + \kappa\cdot\bigl(1.0\cdot\tilde{A}_i - 0.2\cdot\widetilde{\log I_i} - 0.1\cdot E_i\bigr).
\end{align*}
This linear scaling creates stronger dependency effects at high $\kappa$: older age raises risk, while higher income and education are mildly protective. The fixed intercepts ($-1.5$ and $-1.3$) induce a baseline class imbalance favoring \texttt{healthy} outcomes.

\paragraph{Gender.}
Gender is binary with mild dependency on SES, age, and education:
\[
\eta_i = w_{\text{signal}}\cdot(0.3\cdot\text{SES}_i - 0.2\cdot\tilde{A}_i + 0.2\cdot E_i),
\]
\[
G_i \sim \text{Bernoulli}\bigl(\text{logit}^{-1}(\eta_i)\bigr),
\quad G_i\in\{\texttt{female}, \texttt{male}\}.
\]

\paragraph{Controlling dependence strength.}
The global parameter $\kappa\ge 0$ controls the strength of all dependencies:
\begin{itemize}
\item $\kappa=0$: Variables retain weak dependencies due to fixed intercepts in the disease model, but signal contributions vanish ($w_{\text{signal}}=0$).
\item $\kappa=1$ (default): Balanced signal-to-noise ratio, yielding moderate dependencies.
\item $\kappa\gg 1$: Near-deterministic relationships as $w_{\text{signal}}\to 1$ and disease coefficients grow large.
\end{itemize}
Because disease logits scale linearly with $\kappa$ while other variables use the signal-to-noise transformation, disease dependencies strengthen more rapidly at high $\kappa$.

\paragraph{Defaults and realism.}
With default $\kappa=1$, the simulation produces realistic marginal distributions and moderate dependencies: income rises with SES, age, and education; health declines with age but improves with socioeconomic status; and the probabilities of \texttt{diabetic} and \texttt{hypertensive} increase with age and decrease with income and education. These defaults can be adapted to domain-specific baselines by adjusting variable-specific parameters without changing $\kappa$.

\section{Usage example}\label{app:r-code}

RAPID \citep{RAPID} is implemented as an open-source R package with S3 methods for streamlined analysis. A typical workflow proceeds in two stages: first, assess a generated synthetic dataset for per-record risks; second, if evaluating multiple synthesis methods or seeking robust estimates with uncertainty quantification, use cross-validation for method comparison.
The primary use case is evaluating disclosure risk for a specific synthetic dataset before release:

\begin{small}
\begin{verbatim}
# Install and load required packages
devtools::install_github("qwertzlbry/RAPID")
install.packages("synthpop")
library(RAPID)
library(synthpop)

set.seed(2025)
n <- 1000

# Generate independent variables
age <- sample(20:70, n, replace = TRUE)
education <- factor(sample(c("low", "medium", "high"), n, replace = TRUE))
gender <- factor(sample(c("M", "F"), n, replace = TRUE))

# Generate disease_status with dependencies
disease_status <- sapply(1:n, function(i) {
  probs <- c(0.6, 0.2, 0.2)  # Base: healthy, diabetic, hypertensive
  
  # Older people more likely diabetic
  if (age[i] > 55) {
    probs <- c(0.3, 0.5, 0.2)
  }
  
  # Low education + older → more diabetic
  if (age[i] > 55 && education[i] == "low") {
    probs <- c(0.2, 0.6, 0.2)
  }
  
  sample(c("healthy", "diabetic", "hypertensive"), 1, prob = probs)
})
disease_status <- factor(disease_status)

data_orig <- data.frame(age, education, gender, disease_status)
data_syn <- syn(data_orig, method = "cart", seed = 2025)$syn

# Assess attribute inference risk
result <- rapid(
  original_data = data_orig,
  synthetic_data = data_syn,
  quasi_identifiers = c("age", "education", "gender"),
  sensitive_attribute = "disease_status",
  model_type = "rf",
  cat_tau = 0.3,
  return_all_records = TRUE
)

# View result
print(result)

# RAPID Assessment
# ================
# Method: RCS_marginal 
# Risk level: 15.5 %
# Records at risk: 155 / 1000 
# Threshold (tau): 0.3 

# Detailed summary with model metrics
 summary(result)

# RAPID Risk Assessment Summary
# ==============================
# Evaluation method: RCS_marginal 
# Attacker model: Random Forest 
# Risk Metrics:
#   Confidence rate: 0.155 
#   Records at risk: 155 ( 15.5 %)
# Threshold (tau): 0.3 
# Model Performance:
#   Accuracy: 0.599 

# Visualize threshold sensitivity
plot(result)

# Identify high-risk records for disclosure control
print(result, type = "high_risk")

disease_status age edu  gender pred_class true_prob baseline normalized_gain cat_tau at_risk
9    diabetic  68  low    M   diabetic    0.6036335    0.292    0.4401603     0.3    TRUE
14   healthy   42  high   F   healthy     0.6587221    0.495    0.3242022     0.3    TRUE
26   diabetic  63  low    M   diabetic    0.5541064    0.292    0.3702068     0.3    TRUE
... and 152 more

\end{verbatim}
\end{small}

The package provides three S3 methods for result inspection: \texttt{print()} displays a concise summary of risk level and threshold; \texttt{summary()} provides detailed assessment metrics including model performance statistics; and \texttt{plot()} generates threshold sensitivity curves showing how risk estimates vary with $\tau$ (for categorical) or $\varepsilon$ (for continuous attributes). The \texttt{print(result, type = "high\_risk")} call displays records flagged as at-risk, facilitating targeted disclosure control strategies.

When evaluating multiple synthesis methods or seeking robust risk estimates with confidence intervals, cross-validation provides uncertainty quantification. The \texttt{rapid\_synthesizer\_cv()} function implements $k$-fold cross-validation, and results can be inspected using \texttt{print()}:

\begin{small}
\begin{verbatim}

# Define synthesizer function
cart_synthesizer <- function(data, seed = NULL) {
  synthpop::syn(data, method = "cart", m = 1,
                seed = seed, print.flag = FALSE)$syn
}

# Cross-validation across 5 folds
cv_result <- rapid_synthesizer_cv(
  original_data = data_orig,
  synthesizer = cart_synthesizer,
  quasi_identifiers = c("age", "education", "gender"),
  sensitive_attribute = "disease_status",
  k = 5,
  model_type = "rf",
  cat_tau = 0.3
)

print(cv_result)

# RAPID Cross-Validation Results
# ===============================
# Evaluation method: RCS_marginal 
# Attacker model: rf 
# K-folds: 5 
# Threshold (tau): 0.3 
# Threshold (epsilon): NA
# 
# Risk Estimate:
#   Mean: 0.118 
#   SD: 0.03 
#   95% CI: [0.092, 0.144]
\end{verbatim}
\end{small}

Cross-validation is particularly valuable when synthesis methods involve stochastic elements, as it quantifies variability in risk estimates across different training-test splits and enables principled comparison of alternative synthesis approaches.
\end{document}